\documentclass[10pt,twocolumn,letterpaper]{article}

\usepackage[pagenumbers]{cvpr} 

\usepackage{bm}

\definecolor{cvprblue}{rgb}{0.21,0.49,0.74}
\usepackage[pagebackref,breaklinks,colorlinks,allcolors=cvprblue]{hyperref}

\def\paperID{11292} 
\def\confName{CVPR}
\def\confYear{2025}

\title{RSFusionDet: Underwater RGB-Sonar Multimodal Object Detection}

\author{
Zhuoyan Liu$^{1}$\footnotemark[1] \quad Yihan Wang$^{1}$\footnotemark[1] \quad Bo Wang$^{1}$\footnotemark[2] \quad Bing Wang$^{2}$ \quad Ye Li$^{1}$ \\
$^{1}$National Key Laboratory of Autonomous Marine Vehicle Technology, Harbin Engineering University \\
$^{2}$Department of Aeronautical and Aviation Engineering, Hong Kong Polytechnic University \\
{\tt\small \{liuzhuoyan,2020010127,wb\}@hrbeu.edu.cn \quad bingwang@polyu.edu.hk \quad liye@hrbeu.edu.cn}
}

\begin{document}

\maketitle
\begin{abstract}
Underwater unimodal object detection faces many challenges in sensor imaging, such as optical images limited by underwater noise and visible distance, and sonar images limited by less object structural information. While, optical images have rich object structural information, and sonar images are less affected by underwater noise and have a longer visible distance. Optical (RGB modality) and sonar (Sonar modality) images have complementary information underwater. In this paper, we create an RGB-Sonar multimodal object detection dataset, \textbf{R}GB-\textbf{S}onar \textbf{Fusion} (RSFusion) and propose evaluation metrics for the benchmark. And we propose the \textbf{R}GB-\textbf{S}onar \textbf{Fusion} \textbf{Det}ector (RSFusionDet) with a new RGB-Sonar multimodal object detection result expression for RGB-Sonar multimodal object detection. We analyze the features of RGB and Sonar modal information, and design a Cross-Attention Fusion (CAFusion) module to fuse RGB-Sonar spatial misalignment features and Object Matching Head (OMHead) with Loss (OMLoss) to match identical objects in RGB-Sonar modalities. Our RSFusionDet achieves 76.4/48.6 AP (RGB/Sonar) for object detection and 83.4 \(\text{F1-Score}_{match}\) for object matching, on RSFusion, which outperforms other object detection models. Compared with the DINO baseline, our method improves by 0.7/1.4 AP (RGB/Sonar) while simultaneously providing reliable cross-modal object matching. The code and datasets are publicly available at \url{https://github.com/LEFTeyex/RSFusionDet}.
\end{abstract}
\footnotetext[1]{Equal contribution.}
\footnotetext[2]{Corresponding author.}

\section{Introduction}
\label{sec:introduction}

\begin{figure}[t]
\centering

\begin{subfigure}{\linewidth}
\centering
\includegraphics[width=1.0\linewidth]{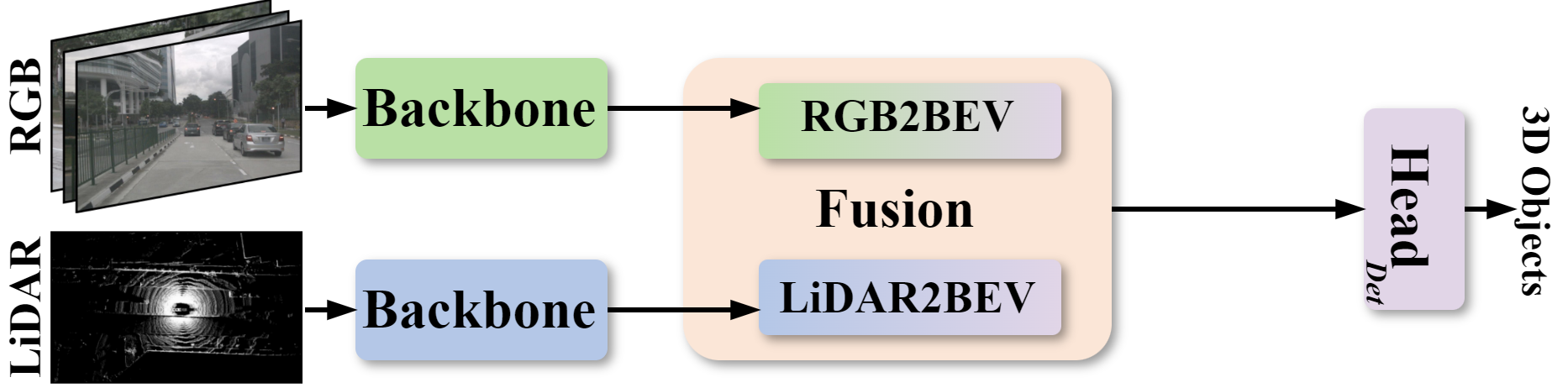}
\caption{Mainstream RGB-LiDAR multimodal object detection.}
\label{fig:rgblidar}
\end{subfigure}
\begin{subfigure}{\linewidth}
\centering
\includegraphics[width=1.0\linewidth]{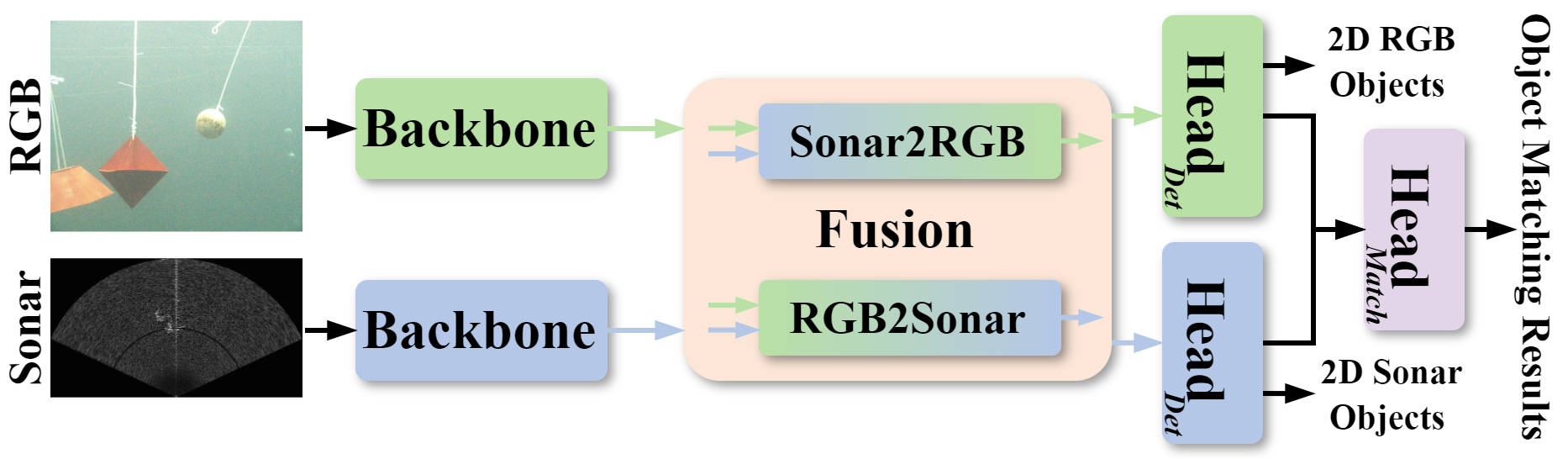}
\caption{RGB-Sonar multimodal object detection (our RSFusionDet).}
\label{fig:rgbsonar}
\end{subfigure}

\caption{Comparisons of multimodal object detection methods with multimodal spatial feature misalignment problem. RGB2BEV represents transforming RGB features to BEV features, the same for LiDAR2BEV. Sonar2RGB represents aligning Sonar features to RGB features, and the same for RGB2Sonar. Differently, our RSFusionDet fuses features and detects objects in the RGB and Sonar feature spaces separately. Then, we use the object matching head to match the identical object in RGB and Sonar modalities.}
\label{fig:abstract}
\end{figure}

Object detection has been developed rapidly in the era of deep learning and is essential for many real-world applications~\cite{fasterrcnn,yolov1,dino,rtdetr}, \textit{e.g.} underwater object detection. Underwater unimodal object detection relies primarily on visual sensors such as optical cameras and sonars (the forward-looking 2D multibeam sonar investigated in this paper). Objects in optical images have rich feature information, including color, texture, shape, edge, \textit{etc}. However, in underwater environments, the absorption and scattering of light by the water medium significantly reduces the visible distance of optical cameras, typically 10-30 meters. In deep or turbid underwater environments, the visible distance may be reduced to 5-10 meters or less. At the same time, optical images also introduce problems such as color cast~\cite{unitmodule} and blur that severely limit the performance and perceived distance of the detector. Objects in sonar images have poor feature information, such as only shape and edge, making it difficult for detectors to identify the object category accurately. However, sonar sensors have a longer visible distance underwater, up to 40-120 meters or more, and are less affected by underwater environmental disturbances. Underwater unimodal object detection using RGB or Sonar modal information alone suffers from various degrees of limitations in perceived distance or detection performance. Optical (RGB modality) and sonar (Sonar modality) images complement each other in terms of feature information and perceived distance, so we investigate the RGB-Sonar multimodal object detection to improve the perceived performance and robustness of underwater object detection.

Recently, RGB-Sonar multimodal perception has attracted increasing attention in underwater robotics. Existing studies mainly focus on RGB-Sonar object tracking, where RGB and sonar images are jointly utilized to improve target localization robustness under challenging underwater environments. Typical representative works include the RGBS50 benchmark and its SCANet tracker~\cite{rgbs50}, which employ cross-attention mechanisms to enhance feature interaction between RGB and sonar modalities for single-object tracking. Compared with RGB-Sonar tracking, RGB-Sonar object detection is a substantially different task. Object detection simultaneously estimates object categories and locations for multiple instances in both modalities, while additionally requiring explicit cross-modal correspondence between detected objects. These requirements introduce several new challenges that have not been addressed in existing RGB-Sonar tracking studies, including spatially misaligned feature fusion for dense scenes, independent object prediction in heterogeneous sensing spaces, and cross-modal object association among multiple detected instances. In this paper, we analyze and study the RGB-Sonar multimodal object detection method. We find that RGB-Sonar multimodal object detection is similar to RGB-LiDAR 3D multimodal object detection in that both have the spatial feature misalignment problem. In the field of autonomous driving, 3D multimodal object detection methods (BEVFusion~\cite{bevfusion}, GAFusion~\cite{gafusion}) utilizing RGB images and LiDAR point clouds achieve RGB-LiDAR spatial misalignment feature fusion by mapping RGB and LiDAR features to the Bird's Eye View (BEV) feature space (using camera and LiDAR calibration parameters and estimated RGB image depth maps). Although the RGB-LiDAR spatial feature misalignment problem has been solved, there are still significant differences from our RGB-Sonar multimodal object detection, their pipelines as shown in Fig. \ref{fig:abstract}. In RGB-LiDAR 3D multimodal object detection, both RGB-LiDAR modalities contain 3D feature information, and RGB-LiDAR multimodal object detection results are uniformly expressed as 3D object detection results. Therefore, RGB-LiDAR features can be uniformly fused in the BEV feature space. In RGB-Sonar multimodal object detection, the Sonar modality contains only 2D feature information and lacks position information in the vertical direction of objects (due to the principle of 2D sonar imaging), and the spatial plane and coverage area of the RGB-Sonar information is slightly different, so RGB-Sonar multimodal object detection results cannot uniformly express as 3D object detection results. It further leads to RGB-Sonar multimodal object detection being unable to perform feature fusion uniformly in the same feature space. In a word, to achieve RGB-Sonar multimodal object detection, the following key problems need to be considered and addressed:

\textbf{(1) How to achieve RGB-Sonar spatial misalignment feature fusion?}
RGB and Sonar images are not only pixel misaligned, but their image data depicts different spatial planes. (In our research, both the camera and sonar are installed facing forward, with RGB images depicting the front view and Sonar images depicting the top view.) Therefore, the features extracted from RGB and Sonar 2D images are spatially misaligned. Since RGB-Sonar multimodal object detection results cannot be expressed uniformly, we choose to fuse RGB-Sonar multimodal features separately in RGB and Sonar feature spaces. We use multi-scale cross-attention to fuse RGB-Sonar multimodal features. There is no definite spatial mapping relationship between RGB and Sonar features, and the scale of the object in RGB and Sonar image features may also be inconsistent. Multi-scale cross-attention can capture global contextual information to construct spatial correspondence between RGB and Sonar features, align the two features, and achieve RGB-Sonar spatial misalignment feature fusion. Meanwhile, multi-scale cross-attention can achieve cross-scale feature fusion of RGB and Sonar to solve the problem of scale inconsistency of objects in RGB and Sonar features. We obtain fused features in the RGB feature space and the Sonar feature space by this feature fusion method, which are used for object detection of RGB and Sonar modalities.

\textbf{(2) How to express RGB-Sonar multimodal object detection results?}
According to the above RGB-Sonar spatial misalignment feature fusion method, we use two detection heads to decode the RGB-Sonar fusion features and express the 2D object detection results in RGB and Sonar modalities, respectively. However, this RGB-Sonar multimodal object detection result lacks the corresponding relationship between identical objects in RGB and Sonar modalities, so it is necessary to perform object matching on object detection results of RGB and Sonar to construct the corresponding relationship between objects in RGB and Sonar modalities. RGB-Sonar multimodal object detection results are expressed using RGB object detection results, Sonar object detection results, and object matching results.

In this paper, we create an underwater RGB-Sonar multimodal object detection benchmark with 7073 RGB-Sonar image pairs and 7 classes to support our research, which is called the \textbf{R}GB-\textbf{S}onar \textbf{Fusion} (\textbf{RSFusion}) dataset. And we also propose evaluation metrics for the underwater RGB-Sonar multimodal object detection. To profit from all this, we present the \textbf{R}GB-\textbf{S}onar \textbf{Fusion} \textbf{Det}ector (\textbf{RSFusionDet}). This dual-branch RGB-Sonar multimodal object detection architecture uses a designed Cross-Attention Fusion module (CAFusion) with deformable attention to align RGB-Sonar features and achieve RGB-Sonar feature fusion. Meanwhile, we design the Object Matching Head (OMHead) and Loss (OMLoss) to match identical objects in RGB and Sonar modalities, and use the object bounding box priori information and IoU weight for the object matching head and loss to improve the performance of object matching. Our method does not require sensor external parameter calibration and feature mapping operations under a fixed sensor configuration, and achieves dynamic fusion of RGB-Sonar features by the global perceptual capacity of attention.

With extensive experiments on the RSFusion dataset, RSFusionDet achieves 76.4/48.6 AP (RGB/Sonar) for object detection and 83.4 \(\text{F1-Score}_{match}\) for object matching. Extensive ablation studies and visualization results validate the effectiveness of our RGB-Sonar feature fusion and object matching methods.

The main contributions are summarized as follows:
\begin{itemize}
    \item We create an underwater RGB-Sonar multimodal object detection benchmark (RSFusion) with 7073 image pairs and propose evaluation metrics for the benchmark to support the research of RGB-Sonar multimodal object detection.
    \item We propose a new dual-branch RGB-Sonar multimodal object detection architecture (RSFusionDet) with a new RGB-Sonar multimodal object detection result expression, achieving RGB-Sonar multimodal object detection.
    \item We design the CAFusion module to fuse RGB-Sonar features, and the OMHead with OMLoss to match identical objects in RGB and Sonar modalities.
    \item Our RSFusionDet is extensively evaluated on the RSFusion dataset, which validates the effectiveness of RGB-Sonar feature fusion and object matching methods, and outperforms other models.
\end{itemize}
\section{Related Work}
\label{sec:related_work}

\subsection{Unimodal Object Detection}
Object detectors based on deep learning are mainly categorized into two-stage and one-stage detectors. Two-stage detectors, such as R-CNN series~\cite{fasterrcnn,sparsercnn}, SPPNet~\cite{sppnet}, RepPoints~\cite{reppoints}, \textit{etc.}, usually use a region proposal network (RPN) to propose priori bounding boxes, which are then refined in the second stage. The most representative one-stage detectors, such as YOLO series~\cite{yolov1,yolov2,yolov3}, YOLOX~\cite{yolox}, SSD~\cite{ssd}, RetinaNet~\cite{retinanet}, FCOS~\cite{fcos}, ATSS~\cite{atss}, TOOD~\cite{tood}, \textit{etc.}, directly output offsets relative to predefined anchors, improving the speed of object detection. In recent years, new fully end-to-end detectors, DETR~\cite{detr}, Deformable DETR~\cite{deformabledetr}, DAB-DETR~\cite{dabdetr}, DINO~\cite{dino}, \textit{etc.}, have emerged, which remove the non-maximum suppression (NMS) from the above detectors. These detectors could also be used for underwater unimodal object detection.

\subsection{Multimodal Object Detection}
Recently, multimodal (multi-sensor) fusion arouses increased interest in object detection, such as RGB-Thermal~\cite{gaff,cssa,cbam,sgfnet,rgbtss,mdnet}, RGB-Event~\cite{rgbe,sodformer}, RGB-LiDAR~\cite{deepfusion,bevfusion,metabev,cmt,gafusion,fullylidar,epnet++}, and RGB-Radar~\cite{rcbevdet,rcmfusion}. GAFF~\cite{gaff} guides RGB-T feature fusion by attention mask. CSSA~\cite{cssa} and Scene-agnostic CBAM~\cite{cbam} utilize channel attention and spatial attention to fuse RGB-T features. The multimodal information fused by RGB-T and RGB-E methods is pixel-aligned, and their object detection results are uniformly expressed as 2D object detection results in the same space. DeepFusion~\cite{deepfusion} makes LiDAR features as queries and constructs aligned RGB features by cross-attention layers for feature fusion. BEVFusion~\cite{bevfusion} utilizes depth estimation of multi-view RGB images and maps RGB features to the BEV feature space based on camera and LiDAR calibration parameters and multi-camera imaging angles. At the same time, the LiDAR features are compressed in the z-axis and finally, the RGB-LiDAR feature fusion is performed by convolution. MetaBEV~\cite{metabev} and CMT~\cite{cmt} construct fused BEV features by learnable BEV queries based on cross-attention layers. GAFusion~\cite{gafusion} introduces sparse depth guidance and LiDAR occupancy guidance to generate 3D features with sufficient depth information to adaptively enhance the interaction of multimodal BEV features from a global perspective. RCM-Fusion~\cite{rcmfusion} uses deformable cross-attention layers and radar-guided BEV queries to achieve feature-level fusion of RGB-Radar modality, and then uses radar grid point refinement to perform their instance-level fusion. RGB-LiDAR and RGB-Radar methods fuse multimodal misalignment features in the BEV feature space, and their object detection results are uniformly expressed as 3D object detection results in the same space. The above methods have achieved superior performance in multimodal object detection. However, they are significantly different from our RGB-Sonar multimodal object detection in terms of multimodal feature fusion and object detection result expression. Their multimodal images are pixel-aligned, or their fused features can be unified in the same feature space. But our RGB-Sonar multimodal images are not pixel-aligned, and our fused features cannot be unified in the same feature space. Their object detection results can be expressed uniformly, but ours cannot. RGB-Sonar multimodal object detection is a new direction to be explored and researched.
\section{RSFusionDet}
\label{sec:method}

\begin{figure*}[t]
\centering
\includegraphics[width=1.0\linewidth]{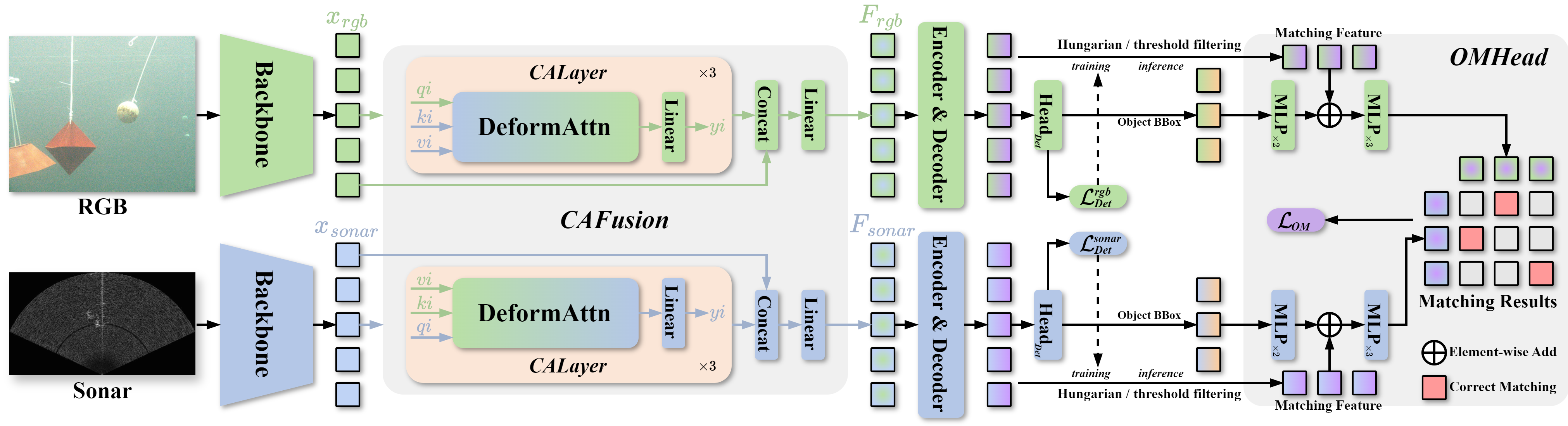}
\caption{The overall architecture of our proposed RSFusionDet. CAFusion aligns multi-scale Sonar features to RGB features using deformable attention (DeformAttn) and fuses the aligned Sonar features and RGB features as RGB fusion features, the same for Sonar fusion features. The fused features are used to calculate the corresponding modal object detection results by encoders, decoders, and heads. OMHead utilizes matching features and object bounding box priori information for RGB-Sonar object matching by cosine similarity, to produce RGB-Sonar multimodal object detection results with object corresponding relationship.}
\label{fig:rsfusiondet}
\end{figure*}

The overall architecture of RSFusionDet is illustrated in Fig. \ref{fig:rsfusiondet}. RGB and Sonar images are fed into two individual and structurally identical backbones to extract features. We fuse multimodal features by cross-attention in both RGB and Sonar feature spaces. RSFusionDet is designed based on the end-to-end DINO~\cite{dino} detector (using the encoder, decoder, and training method of DINO), and its fused RGB and Sonar features are fed into two individual encoder-decoder pipelines to generate RGB and Sonar detection results, respectively. Finally, we apply the cosine similarity, inspired by CLIP~\cite{clip}, to match RGB and Sonar objects based on the output of the decoder to associate identical objects in RGB and Sonar modalities.

\subsection{Cross-Attention Fusion Module}
\label{sec:cafusion}

RGB and Sonar modalities have the spatial feature misalignment problem, and their corresponding object scales may be inconsistent, resulting in scale differences in the multi-scale RGB-Sonar corresponding object features, which is not conducive to RGB-Sonar feature fusion. To address these problems, we propose a Cross-Attention Fusion (CAFusion) module based on the multi-scale deformable attention module~\cite{deformabledetr}, which has a simple architecture and less computational complexity. The CAFusion module contains two pipelines for feature fusion in the RGB and Sonar feature spaces. Each pipeline consists of Cross-Attention Layers (CALayer), where each CALayer consists of a multi-scale deformable attention module and a linear layer followed by a Layer Normalization~\cite{layernorm}.

\begin{figure}[t]
\centering
\includegraphics[width=1.0\linewidth]{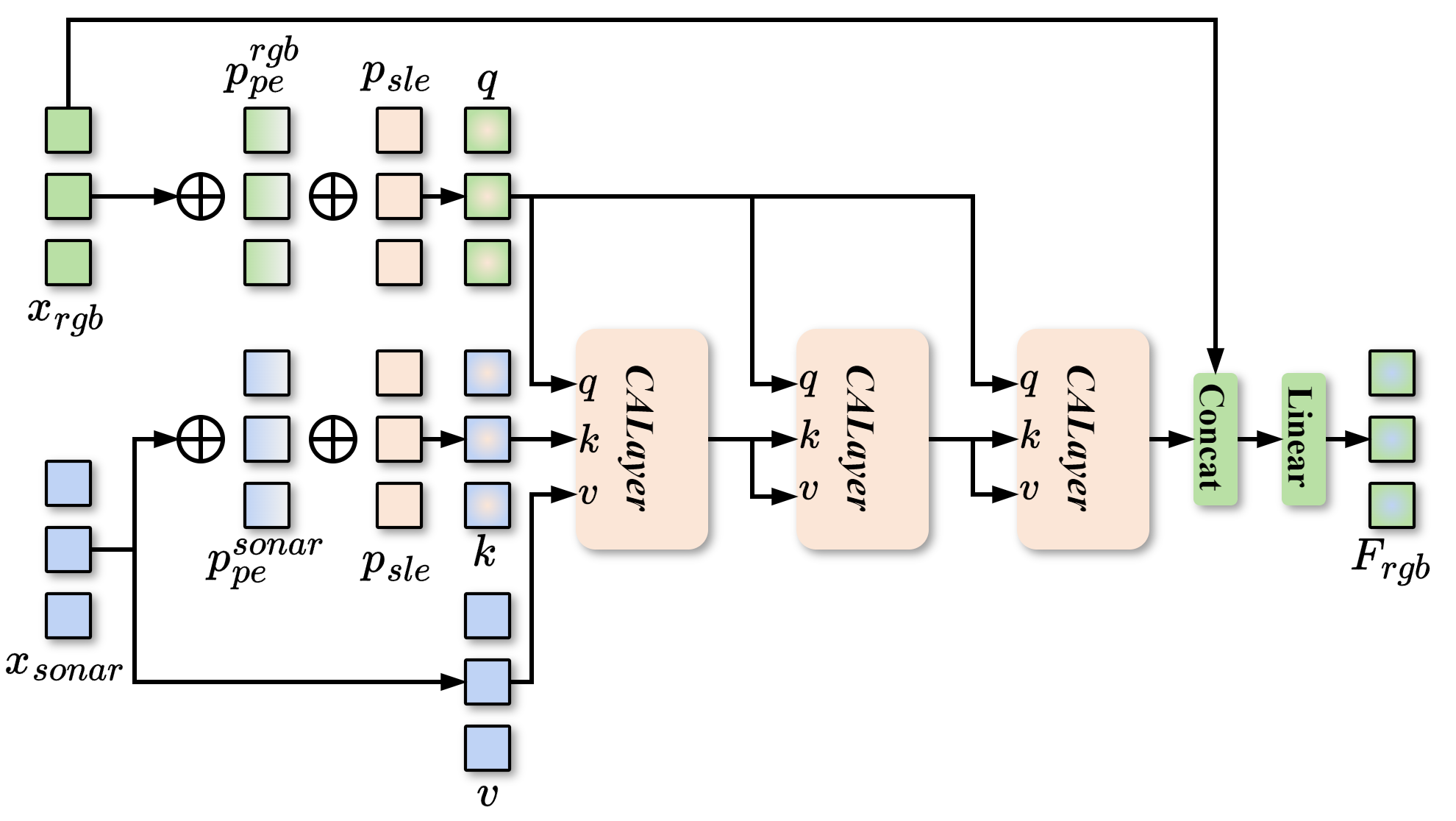}
\caption{The Sonar2RGB fusion part of CAFusion in detail. It aligns Sonar features to RGB features using CALayers and fuses the aligned Sonar features and RGB features, achieving the RGB-Sonar feature fusion in RGB feature space. The RGB2Sonar fusion part is the same as it.}
\label{fig:cafusions2r}
\end{figure}

In the RGB feature space, we construct Sonar features aligned with RGB features and perform feature fusion on them, as in Fig. \ref{fig:cafusions2r}. We use multi-scale RGB features \(\bm{x}_{rgb}\) as query and multi-scale Sonar features \(\bm{x}_{sonar}\) as key and value, and add positional embeddings \(\bm{p}_{pe}\) (spatial positional encoding~\cite{detr}) and learnable scale-level embeddings \(\bm{p}_{sle}\) to the query and key. Making RGB features \(\bm{x}_{rgb}\) as query and Sonar features \(\bm{x}_{sonar}\) as key and value is to query Sonar Value that is of interest to RGB Query to construct aligned Sonar features with the same dimensional size as RGB features. The positional embeddings and scale-level embeddings enable the network to learn the spatial position and scale-level correspondence between RGB and Sonar object features. The output of the CALayer \(\bm{y}\) is used as the key and value for the next CALayer, and the original query is used as the query for the next CALayer to further query sonar features and align them with RGB features. The new key does not add positional embeddings and scale-level embeddings because the spatial position and scale level of the new feature have changed during alignment. The overall process is formulated as follows:
\begin{gather}
\bm{q} = \bm{x}_{rgb}, \quad
\bm{k} = \bm{x}_{sonar}, \quad
\bm{v} = \bm{x}_{sonar} \\
\bm{q} = \bm{q} + \bm{p}_{pe}^{rgb} + \bm{p}_{sle}, \quad
\bm{k} = \bm{k} + \bm{p}_{pe}^{sonar} + \bm{p}_{sle} \\
\bm{y}^{i} = \mathrm{CALayer}^{i} \left( \bm{q}^{i},\bm{k}^{i},\bm{v}^{i} \right) \\
\bm{q}^{i+1} = \bm{q}^{i}, \quad
\bm{k}^{i+1} = \bm{y}^{i}, \quad
\bm{v}^{i+1} = \bm{y}^{i}
\end{gather}
where \(i\) is the index of CALayer, \(\bm{p}_{pe}\) and \(\bm{p}_{sle}\) are added to the input of the first CALayer only (\(\bm{p}_{sle}\) is the same in RGB and Sonar modalities). Then, the aligned Sonar features (the output of the last CALayer) \(\bm{x}_{sonar}^{align}\) constructed by CALayers are concatenated with the original RGB features \(\bm{x}_{rgb}\), and concatenated RGB-Sonar features are fused by a linear layer followed by a Layer Normalization~\cite{layernorm} and a ReLU activation, the RGB-Sonar fused features \(\bm{F}_{rgb}\) in the RGB feature space can be expressed as:
\begin{gather}
\bm{x}_{sonar}^{align} = \bm{y}^{last} \\
\bm{F}_{rgb} = 
\mathrm{Linear}\left(
\mathrm{Concat}(\bm{x}_{rgb}, \bm{x}_{sonar}^{align})
\right)
\label{eq:concat}
\end{gather}
where \(\bm{y}^{last}\) represents the output of the last CALayer, and normalization and activation layers are omitted here. In the Sonar feature space, the operation to get the fused features \(\bm{F}_{sonar}\) is identical to \(\bm{F}_{rgb}\) in the RGB feature space.

Cross-deformable attention can flexibly construct multimodal alignment features by its powerful global perception capability and reduce computational complexity. Meanwhile, multi-scale cross-attention can achieve multimodal cross-scale feature fusion, reducing the effect of scale differences in object features. CAFusion has the advantage of a simple structure and low computational complexity.

\subsection{Object Matching Head and Loss}
\label{sec:omheadloss}

Detection results are independent in RGB and Sonar modalities. To construct the corresponding relationship between identical objects in RGB and Sonar modalities, we design the object matching head and object matching loss.

\begin{figure}[t]
\centering
\includegraphics[width=0.5\linewidth]{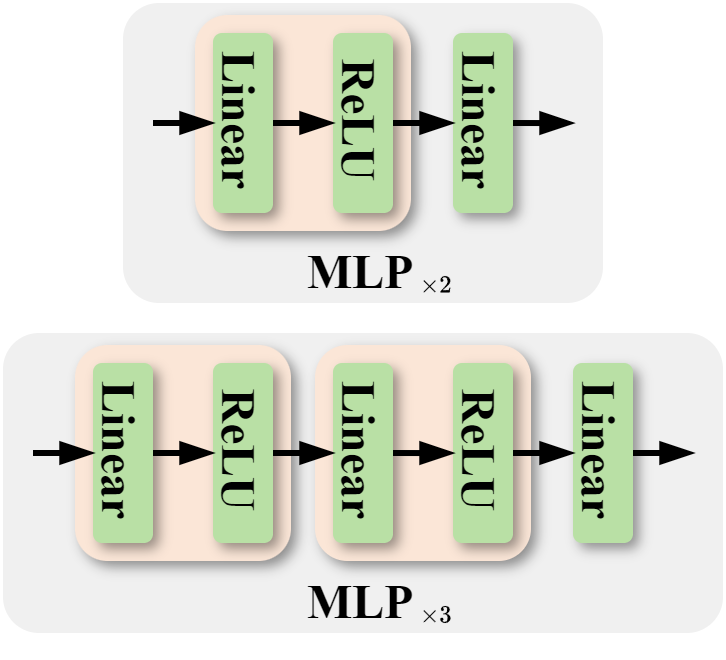}
\caption{The detailed structure of MLP in OMHead.}
\label{fig:mlp}
\end{figure}

\noindent
\textbf{Object Matching Head.}
Object detection and object matching are two complementary but different tasks in RGB-Sonar multimodal perception. While the detector predicts object categories and locations independently in RGB and sonar modalities, it cannot determine whether two detected objects originate from identical objects. Therefore, we introduce the Object Matching Head (OMHead) to explicitly learn semantic correspondence between detected RGB and sonar objects. The primary objective of OMHead is to establish reliable cross-modal object associations. The Object Matching Head calculates the cosine similarity matrix between RGB and Sonar features (we call them matching features) output from the decoder in RGB and Sonar detection pipelines to match objects in RGB and Sonar modalities.

To reduce the computational complexity of object matching, we design two matching feature filtering methods for training and inference to reduce the number of features involved in matching. During training, we use the one-to-one matching results obtained from the Hungarian loss during DETR object detection training to filter features corresponding to GT from the matching features for OMHead. During inference, we set a filtering threshold of 0.3 (according to our RSFusion dataset) for the RGB-Sonar multimodal object detection results and filter the matching features corresponding to the object detection results whose score is greater than the filtering threshold for OMHead.

In addition to the matching features, we also add the object bounding box priori information for object matching to improve the learning of OMHead for the positional relationships of the RGB-Sonar object pair. We encode the object bounding box coordinate (xywh) as a sinusoidal embedding which is the same as the spatial positional encoding in DETR~\cite{detr}, then dynamically encode the embedding using a multi-layer perceptron (MLP), and finally make it the positional embedding and add it to the matching feature.

OMHead uses MLP (the details in Fig. \ref{fig:mlp}) to encode RGB and Sonar matching features separately and calculates the cosine similarity matrix between RGB and Sonar objects. We set an object matching threshold of 0.8 and take RGB-Sonar object pairs with matching scores (cosine similarity scores) greater than the object matching threshold as object matching results. RGB and Sonar objects can only be matched one to one. Once a one-to-many matching result occurs, we take the one with the greatest matching score as the final result. OMHead could not only achieve RGB-Sonar object matching, but also further guide CAFusion to learn the similarity of object features in RGB and Sonar modalities, thereby achieving better RGB-Sonar feature fusion.

\noindent
\textbf{Object Matching Loss.}
Inspired by CLIP~\cite{clip} loss, we develop the Object Matching Loss (OMLoss) \(\mathcal{L}_{OM}\). OMLoss optimizes a symmetric cross-entropy loss to maximize the cosine similarity of identical object embeddings in each pair of RGB-Sonar images and minimize the cosine similarity of non-identical object embeddings for OMHead training.

During training, the OMLoss is computed by the predicted object information which is matched to the ground truth by the Hungarian algorithm. The Hungarian algorithm uses a pair-wise matching cost to match ground truth and predicted objects, but in the initial training stage, the IoU of the predicted object bounding box with its ground truth is small, and the corresponding object features contain less object information. Using these features to train OMHead could lead to a training bias, causing the network to learn a large amount of redundant background information for object matching. To solve this problem and improve the learning efficiency of OMHead, we add an IoU weight that is the product of RGB and Sonar object IoUs between predictions and GTs for OMLoss. In the OMLoss \(\mathcal{L}_{OM}\), the RGB to Sonar object matching loss \(\mathcal{L}_{R2S}\) can be expressed as:
\begin{equation}
\mathcal{L}_{R2S} = - {\textstyle \sum_{i}} \left(
IoU_{i}^{R} * IoU_{i}^{R2S} * \log{}{ ( {e_{i}^{p}} / {{\textstyle \sum_{j}} e_{ij}^{p} } ) }
\right)
\label{eq:r2sloss}
\end{equation}
where \(i\), \(j\) are the index of RGB, Sonar matching objects. \(IoU^{R}\) is the IoU of the predicted RGB object bounding box with its ground truth, and \(IoU^{R2S}\) is the IoU of the Sonar object (matched to the RGB) with its ground truth. \(p\) represents the logits of RGB and Sonar matching objects. The calculation method of Sonar to RGB object matching loss \(\mathcal{L}_{S2R}\) is identical to \(\mathcal{L}_{R2S}\). For RSFusionDet, its total loss \(\mathcal{L}_{RSFusionDet}\) can be expressed as:
\begin{gather}
\mathcal{L}_{OM} = \left( \mathcal{L}_{R2S} + \mathcal{L}_{S2R} \right) / 2 \\
\mathcal{L}_{RSFusionDet} = \mathcal{L}_{Det}^{rgb} + \mathcal{L}_{Det}^{sonar} + \mathcal{L}_{OM}
\label{eq:totalloss}
\end{gather}
where object detection losses \(\mathcal{L}_{Det}^{rgb}\) and \(\mathcal{L}_{Det}^{sonar}\) are the same as the loss which contains the Hungarian loss and the contrastive denoising loss in DINO~\cite{dino}.

\begin{figure}[t]
\centering

\begin{subfigure}{0.28\linewidth}
\centering
\includegraphics[width=1.0\linewidth]{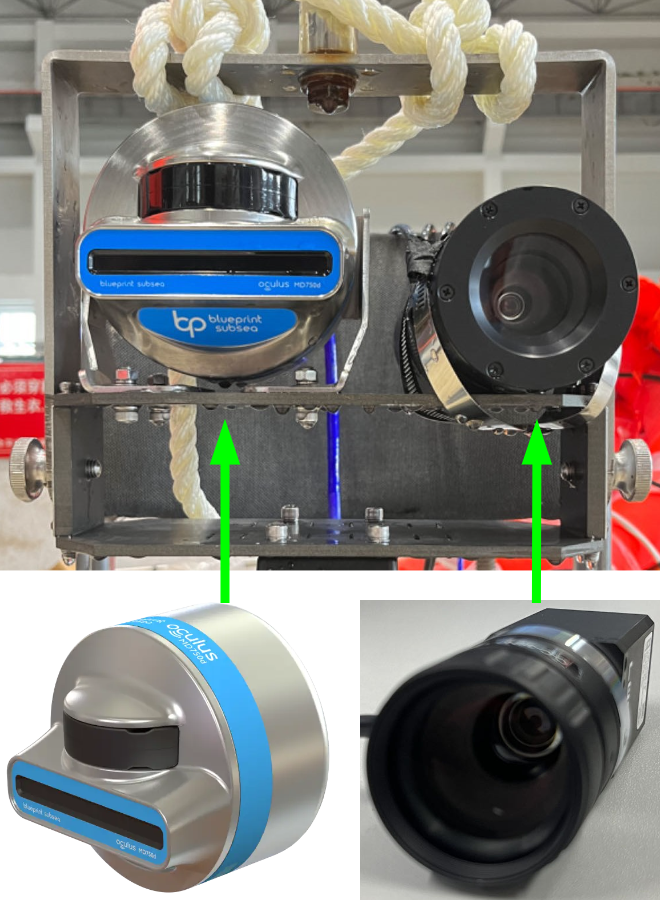}
\caption{Sensors.}
\label{fig:rsfusion_dev}
\end{subfigure}
\hspace{0pt}
\begin{subfigure}{0.68\linewidth}
\centering
\includegraphics[width=1.0\linewidth]{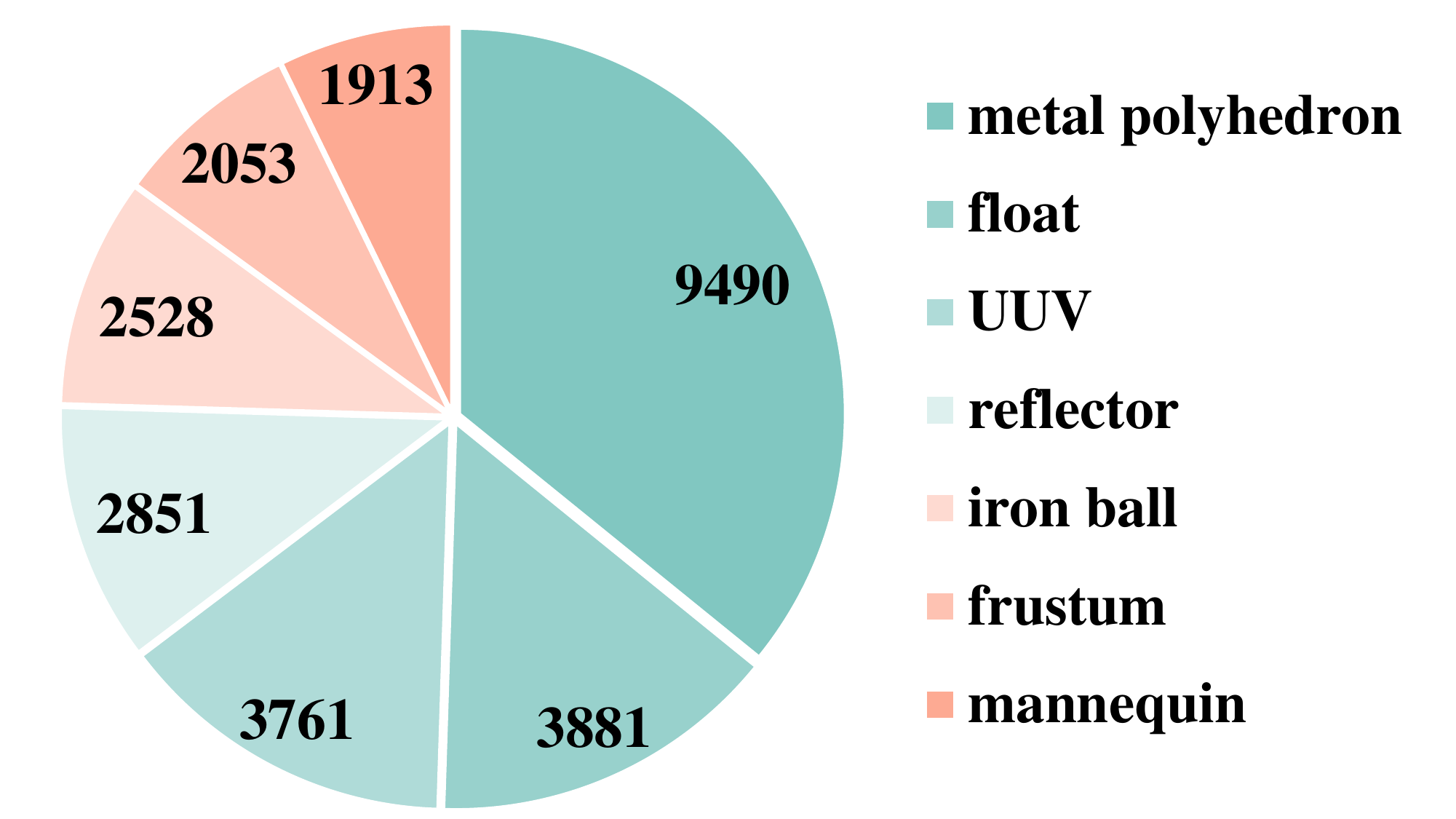}
\caption{Distribution of different object quantities.}
\label{fig:rsfusion_cls}
\end{subfigure}

\caption{A brief introduction to our RSFusion dataset.}
\label{fig:RSFusion}
\end{figure}

\section{Datasets and Evaluation Metrics}
\label{sec:dataset_evaluation}

\subsection{RSFusion Dataset}
\label{sec:dataset}

We conduct experiments on the underwater RGB-Sonar multimodal object detection dataset, RGB-Sonar Fusion (RSFusion) we create, which includes 7 classes, mannequin, UUV, reflector, metal polyhedron, float, iron ball, and frustum. We use a LUCID TRI050S-QC camera (in Fig. \ref{fig:rsfusion_dev} right) and an Oculus MD750d 2D sonar (in Fig. \ref{fig:rsfusion_dev} left) with a 1.2 MHz operating frequency to simultaneously capture paired RGB-Sonar images in a tank. The RSFusion dataset includes paired images in light and dark scenes, and paired images at maximum sonar ranges of 5, 10, 15, and 20 meters. The maximum sonar range represents the maximum range at which the sonar sensor scans and images, which can be set. The scanning range is centered around the sonar and a sector area with a radius of the set parameters. The horizontal and vertical fields of view (FOV) for the camera are 30.75 and 23.31 degrees, respectively, while the horizontal FOV for the sonar is 130 degrees. The dataset annotation also includes metadata such as the number of beams \textit{n\_beams}, the corresponding beam angle \textit{beam\_angles}, the range resolution \textit{range\_resolution}, and so on in each sonar image and the corresponding relationship flag of identical objects in RGB and Sonar modalities. The RSFusion dataset is split into 5658/1415 RGB-Sonar image pairs for training/validation (the following data is presented in this format), including 4330/1080 and 1328/335 image pairs in light and dark scenes. We also count the number of paired images when the maximum sonar range is 5, 10, 15, and 20 meters, which are 1559/388, 593/147, 2934/729, and 572/151 image pairs, respectively. The RSFusion dataset contains 21202/5275 objects for training/validation of both RGB and Sonar images, including 4084/1032 object pairs in RGB and Sonar modalities. The distribution of objects in the dataset is shown in Fig. \ref{fig:rsfusion_cls}. RSFusion also contains some image pairs where only RGB or Sonar has no objects, and these data are used to improve the robustness of the model for feature fusion. The RGB image size is 1024\(\times\)1224. The Sonar image has different sizes, 374\(\times\)678-377\(\times\)684 and 499\(\times\)905-505\(\times\)916, which are related to the maximum sonar range.

\subsection{Evaluation Metrics}
\label{sec:evaluation}

The evaluation metrics of RGB-Sonar multimodal object detection include two parts, object detection metrics and object matching metrics.

\subsubsection{Object Detection Metrics}
\label{sec:odm}

For the object detection metrics, we follow the standard COCO~\cite{coco} evaluation protocol and report the Average Precision (\(\text{AP}\)) for RGB and Sonar object detection, respectively. Furthermore, according to different kinds of data in the RSFusion dataset, we set the metrics for underwater light and dark scenes (\(\text{AP}_{light}, \text{AP}_{dark}\)) to evaluate the complementarity between modalities and their robustness to different scenes and the metrics for Sonar modality in different maximum sonar range settings (\(\text{AP}_{sr5}, \text{AP}_{sr10}, \text{AP}_{sr15}, \text{AP}_{sr20}\)) to evaluate the impact of different maximum sonar ranges on the performance of RGB-Sonar feature fusion. All evaluation metrics for both RGB and Sonar object detection are as follows:

\medskip
\noindent\(\text{AP}\) (\(\text{AP}_{50:95}\)): Average AP value at the IoU threshold from 0.5 to 0.95 (step size 0.05).

\noindent\(\text{AP}_{50}\): AP value at the IoU threshold is 0.5.

\noindent\(\text{AP}_{75}\): AP value at the IoU threshold is 0.75.

\noindent\(\text{AP}_{s}\): AP value for small objects.

\noindent\(\text{AP}_{m}\): AP value for medium objects.

\noindent\(\text{AP}_{l}\): AP value for large objects.

\noindent\(\text{AP}_{light}\): \(\text{AP}_{50:95}\) value for objects in the light scene.

\noindent\(\text{AP}_{dark}\): \(\text{AP}_{50:95}\) value for objects in the dark scene.

\noindent\(\text{AP}_{sr5}\): \(\text{AP}_{50:95}\) value for objects at the sr is 5 meters.

\noindent\(\text{AP}_{sr10}\): \(\text{AP}_{50:95}\) value for objects at the sr is 10 meters.

\noindent\(\text{AP}_{sr15}\): \(\text{AP}_{50:95}\) value for objects at the sr is 15 meters.

\noindent\(\text{AP}_{sr20}\): \(\text{AP}_{50:95}\) value for objects at the sr is 20 meters.

\subsubsection{Object Matching Metrics}
\label{sec:omm}

For RGB-Sonar object matching, we design evaluation metrics (\(\text{P}_{match}\), \(\text{R}_{match}\), and \(\text{F1-Score}_{match}\)) that is almost identical to Precision, Recall, and F1-Score in object detection. We assign the object with the maximum IoU to the bounding box ground truth by calculating the IoU between the object and the ground truth in the predicted matching results (and remove the result with IoU less than 0), and denote the number of assigned matching object pairs as the total number of samples predicted as correct matches \(N_{pred}\). Meanwhile, we calculate the number of correctly matched samples \(N\) in \(N_{pred}\), and denote the total number of samples of the object matching ground truth as \(N_{gt}\). The evaluation metrics for object matching can be expressed as:
\begin{gather}
\text{P}_{match} = N / N_{pred}, \quad \text{R}_{match} = N / N_{gt} \\
\text{F1-Score}_{match} = 2 * \frac{\text{P}_{match} * \text{R}_{match}}{\text{P}_{match} + \text{R}_{match}}
\end{gather}
All evaluation metrics for RGB-Sonar object matching are as follows:

\medskip
\noindent\(\text{P}_{match}\): Precision for RGB-Sonar object matching.

\noindent\(\text{R}_{match}\): Recall for RGB-Sonar object matching.

\noindent\(\text{F1-Score}_{match}\): F1-Score for RGB-Sonar object matching.
\section{Experiments}
\label{sec:experiments}

\begin{figure*}[t]
\centering
\includegraphics[width=1.0\linewidth]{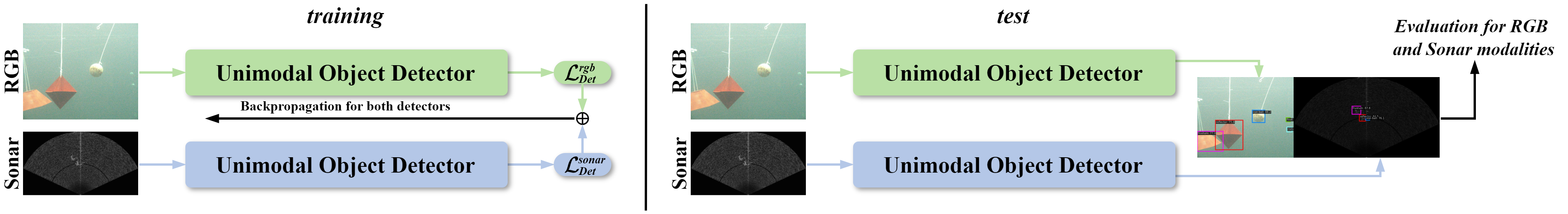}
\caption{RGB-Sonar separate object detection framework for training and test.}
\label{fig:rgbs_sep_stream}
\end{figure*}
\begin{table*}[t]
\caption{Comparisons with other state-of-the-art models. The metrics are object detection APs in RGB/Sonar modality. Our RSFusionDet has object matching evaluation metrics (80.9 \(\text{P}_{match}\), 86.0 \(\text{R}_{match}\), and 83.4 \(\text{F1-Score}_{match}\)) in Table \ref{tab:ab_ofm_threshold}. Other models lack object matching, so their evaluation metrics are ignored here. \dag\space denotes the adoption of its fusion method.}
\centering
\resizebox{\linewidth}{!}{

\begin{tabular}{@{}lccccccccc@{}}
\toprule
\textbf{Model} &
\textbf{Modality} &
\textbf{\#Params (M)} &
\textbf{GFLOPs} &
\textbf{AP} &
\(\textbf{AP}_{50}\) &
\(\textbf{AP}_{75}\) &
\(\textbf{AP}_{s}\) &
\(\textbf{AP}_{m}\) &
\(\textbf{AP}_{l}\) \\ \midrule
Faster R-CNN~\cite{fasterrcnn}          & R/S & 83  & 179 & 73.4 / 42.9 & 97.0 / 86.4 & 82.7 / 38.3 & 15.4 / 37.3 & 38.7 / 47.0 & 79.5 / 68.7 \\
YOLOv3~\cite{yolov3}                    & R/S & 123 & 152 & 66.2 / 38.4 & 93.5 / 81.5 & 74.6 / 31.1 & 12.9 / 32.4 & 33.0 / 42.1 & 72.9 / 65.5 \\
FCOS~\cite{fcos}                        & R/S & 64  & 154 & 70.8 / 39.3 & 96.6 / 83.2 & 79.7 / 31.9 & 12.4 / 33.8 & 35.5 / 43.1 & 77.4 / 68.1 \\
ATSS~\cite{atss}                        & R/S & 64  & 158 & 73.6 / 43.5 & 96.6 / 85.7 & 82.4 / 38.9 & 11.6 / 38.3 & 35.2 / 46.8 & 80.0 / 69.1 \\
Sparse R-CNN~\cite{sparsercnn}          & R/S & 212 & 129 & 70.3 / 40.4 & 95.1 / 86.6 & 78.1 / 30.1 & 10.8 / 36.6 & 35.0 / 42.4 & 78.1 / 67.7 \\
TOOD~\cite{tood}                        & R/S & 64  & 155 & 63.3 / 44.7 & 89.5 / 86.5 & 71.0 / 40.7 & 10.2 / 38.6 & 28.8 / 47.4 & 68.9 / 67.4 \\
YOLOX-L~\cite{yolox}                    & R/S & 108 & 152 & 68.4 / 43.0 & 90.9 / 89.8 & 76.3 / 36.4 &  9.3 / 37.0 & 32.5 / 45.7 & 77.0 / 67.1 \\
Deformable DETR++~\cite{deformabledetr} & R/S & 82  & 160 & 72.1 / 43.5 & 97.0 / 91.7 & 80.4 / 35.2 & 14.2 / 38.7 & 33.5 / 46.1 & 78.6 / 67.8 \\
DAB-DETR~\cite{dabdetr}                 & R/S & 87  & 85  & 63.5 / 30.8 & 94.0 / 73.4 & 71.4 / 18.5 & 22.7 / 25.7 & 26.7 / 35.0 & 71.0 / 57.8 \\
DINO~\cite{dino}                        & R/S & 95  & 233 & 75.7 / 47.2 & 97.1 / 93.7 & 83.0 / 41.5 & 21.5 / 42.7 & 45.3 / 51.4 & 83.2 / 69.8 \\ \midrule
DINO+GAFF\dag~\cite{gaff}               & R+S & 96  & 240 & 75.9 / 47.8 & 97.3 / 94.0 & 83.2 / 42.2 & 22.5 / 43.1 & 48.0 / 52.5 & 83.6 / 70.0 \\
DINO+SCANet\dag~\cite{rgbs50}           & R+S & 97  & 245 & 76.1 / 48.1 & 97.4 / 94.2 & 83.3 / 42.8 & 22.9 / 43.4 & 49.2 / 53.0 & 83.7 / 70.1 \\ \midrule
RSFusionDet                             & R+S & 98  & 249 &
                                        \textbf{76.4} / \textbf{48.6} &
                                        \textbf{97.5} / \textbf{94.4} &
                                        \textbf{83.5} / \textbf{43.9} &
                                        \textbf{23.6} / \textbf{43.8} &
                                        \textbf{51.1} / \textbf{54.1} &
                                        \textbf{83.9} / \textbf{70.3} \\ \bottomrule
\end{tabular}

}
\label{tab:com_all}
\end{table*}
\begin{table}[t]
\caption{Comparisons with other state-of-the-art models with CAFusion and OMHead. CA and OM represent CAFusion and OMHead, respectively. And RSFusionDet is \(\text{DINO}_\textbf{+CA+OM}\).}
\centering
\resizebox{\linewidth}{!}{

\begin{tabular}{@{}lcc@{}}
\toprule
\textbf{Model} &
\textbf{Modality} &
\textbf{AP} \\ \midrule
\(\text{Faster R-CNN}_\textbf{+CA+OM}\)             & R+S & 74.2(+0.8) / 44.3(+1.4) \\
\(\text{YOLOv3}_\textbf{+CA+OM}\)                   & R+S & 66.8(+0.6) / 39.5(+1.1) \\
\(\text{FCOS}_\textbf{+CA+OM}\)                     & R+S & 71.5(+0.7) / 40.5(+1.2) \\
\(\text{ATSS}_\textbf{+CA+OM}\)                     & R+S & 74.3(+0.7) / 44.8(+1.3) \\
\(\text{Sparse R-CNN}_\textbf{+CA+OM}\)             & R+S & 71.0(+0.7) / 41.6(+1.2) \\
\(\text{TOOD}_\textbf{+CA+OM}\)                     & R+S & 63.9(+0.6) / 46.2(+1.5) \\
\(\text{YOLOX-L}_\textbf{+CA+OM}\)                  & R+S & 69.0(+0.6) / 44.3(+1.3) \\
\(\text{Deformable DETR++}_\textbf{+CA+OM}\)        & R+S & 72.8(+0.7) / 44.8(+1.3) \\
\(\text{DAB-DETR}_\textbf{+CA+OM}\)                 & R+S & 64.1(+0.6) / 31.8(+1.0) \\ \midrule
RSFusionDet (\(\text{DINO}_\textbf{+CA+OM}\))       & R+S & \textbf{76.4(+0.7)} / \textbf{48.6(+1.4)} \\ \bottomrule
\end{tabular}

}
\label{tab:com_all_caom}
\end{table}

\subsection{Implementation Details}
\label{sec:id}

All models are implemented and tested based on the MMDetection~\cite{mmdetection} framework. We adopt the pre-trained ResNet-50~\cite{resnet} as the backbone for RGB and Sonar stream for RSFusionDet. All models are trained on 2 NVIDIA A6000 GPUs with batch size 4 per GPU. The RGB and Sonar input image resolutions are 512\(\times\)640 and 512\(\times\)928, respectively, and we use the horizontal RandomFlip data augmentation on the input image. During the training stage, we use AdamW~\cite{adamw} optimizer, with weight decay of \(1e^{-4}\) and initial learning rates (lr) as \(1e^{-4}\) and \(1e^{-5}\) for the encoder-decoder and backbone, respectively. We take overall 12 training epochs for RSFusionDet and drop lr at the 11th epoch multiplying by 0.1.

In the comparison experiment, both RGB-LiDAR methods and our RGB-Sonar method have the spatial feature misalignment problem. However, these RGB-LiDAR fusion methods~\cite{bevfusion,metabev,cmt,gafusion}, cannot be applied to our RGB-Sonar fusion method. Most of these methods require feature mapping using predicted depth maps and sensor calibration parameters, or require priori object coordinate position information, which cannot be implemented in our RGB-Sonar multimodal object detection method.

During training and test, both the RGB and Sonar images of RSFusionDet are inputted in pairs simultaneously. In order to achieve fairness in comparison with unimodal object detection methods, we design a dual-stream RGB-Sonar separate object detection framework, ensuring consistency between the RGB and Sonar image data streams of the unimodal object detection method and our RSFusionDet during training and test. As shown in Fig. \ref{fig:rgbs_sep_stream}, the RGB-Sonar separate object detection framework we design includes two independent unimodal object detectors for RGB and Sonar modalities, respectively. Like our RSFusionDet, it can simultaneously train and test for both RGB and Sonar modalities. During the training, the losses of these two unimodal object detectors are summed up to train them for both RGB and Sonar modalities simultaneously, making them the same as our RSFusionDet training data stream for fair comparison experiments. Difference from RSFusionDet, the RGB-Sonar separate object detection framework cannot implement RGB-Sonar feature fusion and RGB-Sonar object matching. In the experiment, we train and test unimodal object detection models on the RSFusion dataset using the RGB-Sonar separate object detection framework without RGB-Sonar feature fusion, and call its modality as RGB or Sonar, \textit{i.e.} R/S. Meanwhile, we call the modality of RSFusionDet with RGB-Sonar feature fusion as the fusion of RGB and Sonar, \textit{i.e.} R+S. In the experiment, all models are trained and tested on the training set and the validation set of the RSFusion dataset.

Furthermore, at present, the only publicly available RGB-Sonar related dataset is RGBS50~\cite{rgbs50}. This dataset is an RGB-Sonar single object tracking dataset constructed in our laboratory, which consists of continuous frame images with single object annotation information. Its object diversity is far less than our RSFusion dataset. And both RGBS50 and RSFusion datasets are compiled from the same experimental data, so it is convincing enough to conduct validation only on the RSFusion dataset. Although both RGBS50 and RSFusion datasets are compiled from the same experimental data, every frame is independently re-annotated for multimodal object detection. Compared with RGBS50, which only requires tracking one initialized object, RSFusion requires annotating all visible objects in both RGB and sonar images, assigning category labels, generating modality-specific bounding boxes, and establishing cross-modal correspondence labels for matched object pairs.

\begin{figure*}[t]
\centering
\includegraphics[width=1.0\linewidth]{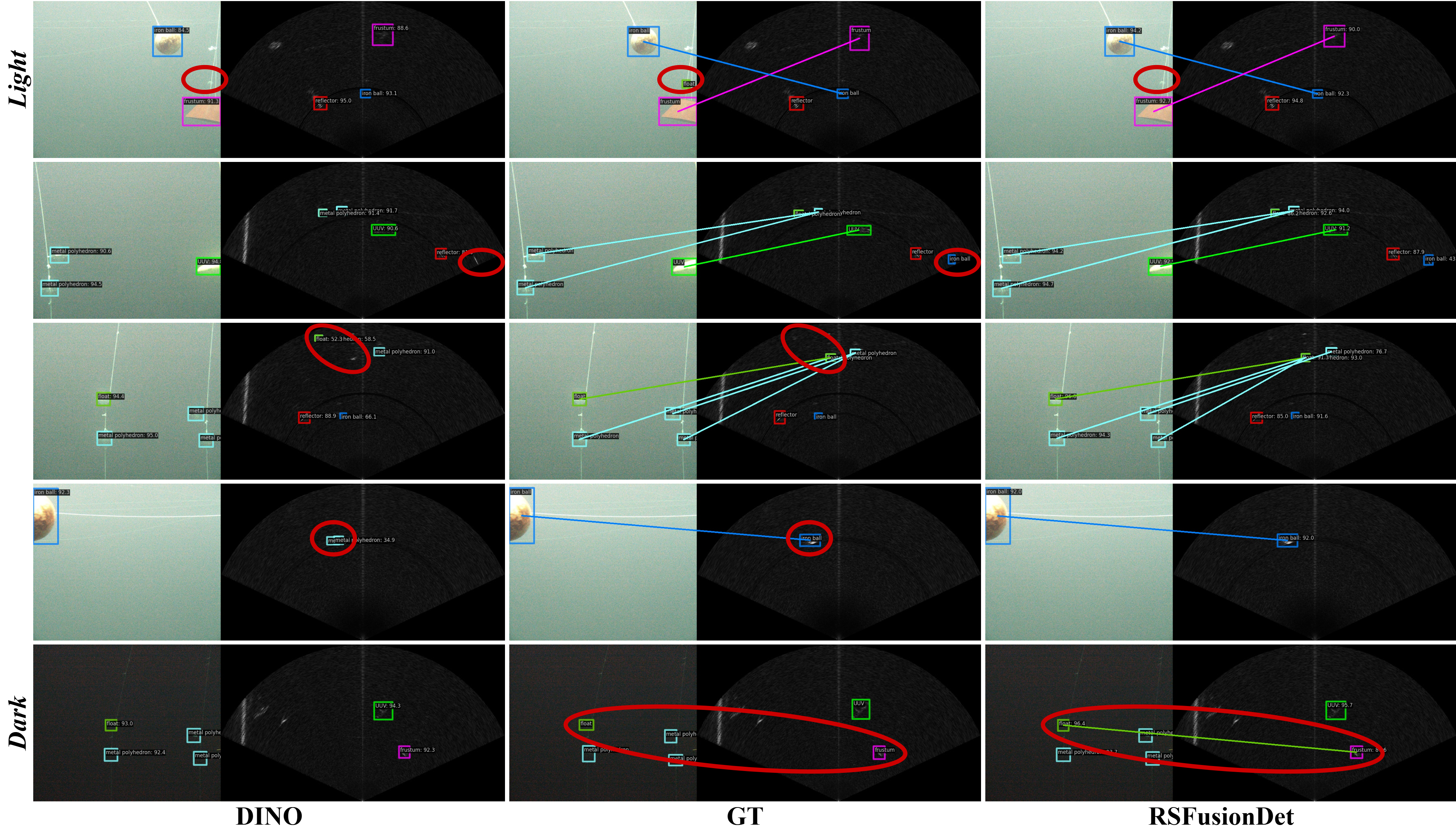}
\caption{Comparisons of object detection and matching results (lines) on the RSFusion validation set with light and dark scenes. Note that some objects appear only in one modality because the RGB camera and forward-looking sonar have different fields of view and sensing ranges. Such modality-specific observations are common in practical underwater perception systems.}
\label{fig:detection}
\end{figure*}
\begin{figure*}[t]
\centering
\includegraphics[width=1.0\linewidth]{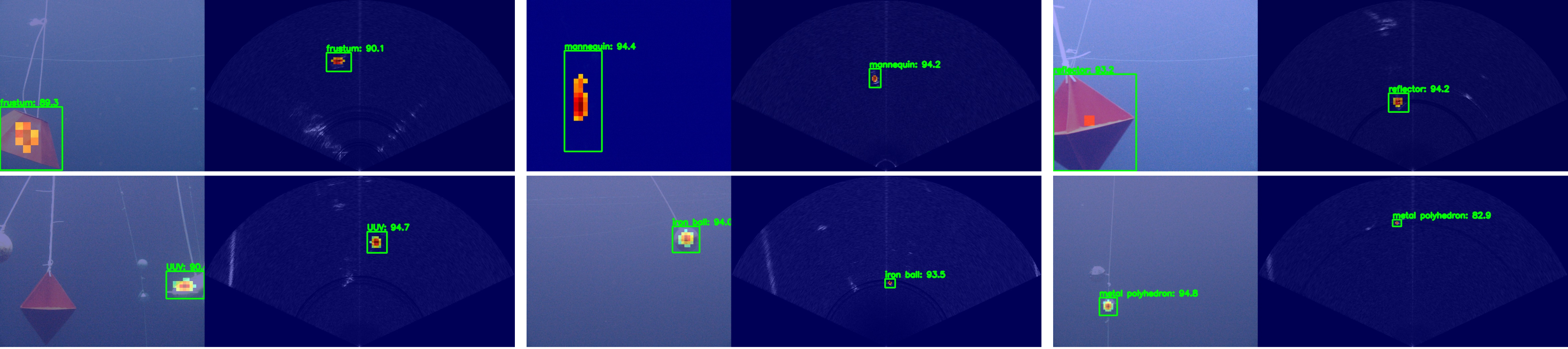}
\caption{RGB-Sonar corresponding object attention maps by Grad-CAM~\cite{gradcam} method. The attention map is visualized on the layer behind the CAFusion module. The RGB-Sonar object attention gradient is obtained only from the object \textit{bbox} and \textit{iou} losses in RGB modality. The attention in RGB and Sonar images represents the degree of correlation between object features in RGB and Sonar modalities, indicating the feature fusion effect of the CAFusion module.}
\label{fig:attention}
\end{figure*}
\begin{table*}[t]
\caption{Comparisons (AP) on different classes in RSFusion dataset.}
\centering
\resizebox{\linewidth}{!}{

\begin{tabular}{@{}lcccccccc@{}}
\toprule
\textbf{Model} &
\textbf{Modality} &
\textbf{mannequin} &
\textbf{UUV} &
\textbf{reflector} &
\textbf{metal polyhedron} &
\textbf{float} &
\textbf{iron ball} &
\textbf{frustum} \\ \midrule
Faster R-CNN      & R/S & 78.2 / 60.1 & 76.7 / 54.2 & 86.2 / 58.4 & 51.0 / 23.6 & 66.0 / 21.9 & 75.1 / 36.7 & 80.4 / 45.6 \\
YOLOv3            & R/S & 60.6 / 54.8 & 66.4 / 48.8 & 80.5 / 51.6 & 50.9 / 17.8 & 62.1 / 22.8 & 70.1 / 34.3 & 72.5 / 38.5 \\
FCOS              & R/S & 72.6 / 54.9 & 72.8 / 53.6 & 82.3 / 55.2 & 51.5 / 18.1 & 64.3 / 19.7 & 74.5 / 32.8 & 77.6 / 40.9 \\
ATSS              & R/S & 79.2 / 60.6 & 73.5 / 54.4 & 86.3 / 59.7 & 51.5 / 24.3 & 65.5 / 18.8 & 78.8 / 38.6 & 80.2 / 48.3 \\
Sparse R-CNN      & R/S & 75.1 / 56.6 & 75.0 / 51.1 & 82.3 / 51.6 & 47.4 / 22.3 & 63.9 / 29.2 & 74.3 / 30.6 & 74.0 / 41.3 \\
TOOD              & R/S & 54.0 / 61.0 & 62.2 / 54.5 & 74.5 / 60.0 & 51.7 / 26.6 & 65.6 / 22.4 & 68.1 / 40.4 & 66.7 / 47.8 \\
YOLOX-L           & R/S & 65.7 / 57.6 & 77.4 / 51.8 & 83.2 / 56.2 & 48.6 / 24.5 & 59.2 / 27.1 & 68.9 / 37.9 & 75.9 / 45.9 \\
Deformable DETR++ & R/S & 79.4 / 59.0 & 77.3 / 49.8 & 83.7 / 53.7 & 51.0 / 31.5 & 64.4 / 31.2 & 72.1 / 35.6 & 76.7 / 43.5 \\
DAB-DETR          & R/S & 62.2 / 42.9 & 63.0 / 39.7 & 79.9 / 42.7 & 43.9 / 16.7 & 60.6 / 11.3 & 65.2 / 24.7 & 69.9 / 37.5 \\
DINO              & R/S & 84.0 / 61.9 & 81.5 / 54.5 & 88.1 / 58.6 & 50.8 / 35.7 & 66.0 / 33.6 & 79.0 / 39.9 & 80.7 / 46.3 \\ \midrule
DINO+GAFF\dag     & R+S & 84.1 / 62.0 & 81.7 / 54.6 & 88.3 / 59.1 & 51.2 / 36.0 & 66.0 / 34.5 & 79.1 / 40.2 & 81.3 / 47.1 \\
DINO+SCANet\dag   & R+S & 84.3 / 62.0 & 81.8 / 54.7 & 88.4 / 59.6 & 51.6 / 36.3 & 66.0 / 35.4 & 79.2 / 40.5 & 82.0 / 47.9 \\ \midrule
RSFusionDet       & R+S &
                  \textbf{84.4} / \textbf{62.1} &
                  \textbf{82.0} / \textbf{54.8} &
                  \textbf{88.6} / \textbf{60.1} &
                  \textbf{52.0} / \textbf{36.7} &
                  \textbf{66.0} / \textbf{36.5} &
                  \textbf{79.3} / \textbf{40.9} &
                  \textbf{82.7} / \textbf{48.8} \\ \bottomrule
\end{tabular}

}
\label{tab:com_cls}
\end{table*}
\begin{table}[t]
\caption{Comparisons (AP) on light and dark scenes.}
\centering
\resizebox{\linewidth}{!}{

\begin{tabular}{@{}lccc@{}}
\toprule
\textbf{Model} &
\textbf{Modality} &
\(\textbf{AP}_{light}\) &
\(\textbf{AP}_{dark}\) \\ \midrule
Faster R-CNN      & R/S & 76.7 / 45.0 & 58.5 / 33.4 \\
YOLOv3            & R/S & 68.2 / 40.8 & 52.0 / 28.5 \\
FCOS              & R/S & 74.5 / 41.0 & 56.4 / 30.4 \\
ATSS              & R/S & 76.5 / 45.0 & 59.0 / 34.5 \\
Sparse R-CNN      & R/S & 73.9 / 42.3 & 55.5 / 33.8 \\
TOOD              & R/S & 67.2 / 46.6 & 41.2 / 34.0 \\
YOLOX-L           & R/S & 74.5 / 44.5 & 43.5 / 35.7 \\
Deformable DETR++ & R/S & 75.5 / 45.0 & 58.8 / 36.4 \\
DAB-DETR          & R/S & 66.6 / 32.4 & 49.0 / 25.4 \\
DINO              & R/S & 79.8 / 48.8 & 59.6 / 40.3 \\ \midrule
DINO+GAFF\dag     & R+S & 80.0 / 49.2 & 59.9 / 40.7 \\
DINO+SCANet\dag   & R+S & 80.3 / 49.5 & 60.2 / 41.1 \\ \midrule
RSFusionDet       & R+S &
                  \textbf{80.6} / \textbf{49.9} &
                  \textbf{60.6} / \textbf{41.5} \\ \bottomrule
\end{tabular}

}
\label{tab:com_sce}
\end{table}

\subsection{Comparisons with State-of-the-arts}
\label{sec:cs}

\subsubsection{Comparisons with Other Models}

As shown in Table \ref{tab:com_all}, we report the results of RSFusionDet on the RSFusion validation set, and compare them with other state-of-the-art unimodal models without RGB-Sonar feature fusion and object matching. The results show that RSFusionDet outperforms all other models, with 76.4{\small (+0.7)}/48.6{\small (+1.4)} AP (RGB/Sonar AP) (compared to the best model DINO~\cite{dino}). RSFusionDet significantly improves both RGB and Sonar detection, as RGB provides richer feature information for Sonar, resulting in higher improvement on detection performance in the Sonar modality. And RSFusionDet has a significant performance improvement across all object scales and outperforms all other models, especially on medium objects with a 51.1/54.1 AP. It proves that RGB-Sonar feature fusion is beneficial for improving the performance of object detection. Compared to its base model DINO, RSFusionDet has added CAFusion and OMHead, but only gains 3 M \#Params and 16 GFLOPs. Meanwhile, we plug CAFusion and OMHead into other unimodal models and compare them with our RSFusionDet in Table \ref{tab:com_all_caom} (Note that RSFusionDet is equivalent to DINO with CAFusion and OMHead.). After plugging in CAFusion and OMHead, these models can also achieve RGB-Sonar feature fusion, with improvements of 0.6-0.8 AP and 1.0-1.5 AP in RGB and Sonar modalities, respectively. And our RSFusionDet still outperforms other models with CAFusion and OMHead. It further proves the effectiveness of CAFusion and OMHead in RGB-Sonar feature fusion. Besides, we visualize the detection results and the RGB-Sonar object attention maps of RSFusionDet, as in Figs. \ref{fig:detection} and \ref{fig:attention}. DINO has many erroneous detections and omissions, while our RSFusionDet has achieved more accurate detection results in Fig. \ref{fig:detection}. Moreover, our RSFusionDet enables effective RGB-Sonar object matching. RSFusionDet achieves more accurate object localization and classification in Sonar modality, which compensates for the poor accuracy of object classification in Sonar modality. It proves that RGB-Sonar multimodal object detection is beneficial for improving the detection performance between modalities. However, RSFusionDet faces difficulties in detecting small objects with occlusion (in the first row of Fig. \ref{fig:detection}), mainly due to the influence of underwater noise (the attenuation effect of the water medium on light waves). And RSFusionDet also has the problem of RGB-Sonar object mismatch (the green line connecting the float in RGB modality and the frustum in Sonar modality), as shown in the last row of Fig. \ref{fig:detection}. It is mainly caused by the high similarity of object features between different categories of RGB and Sonar modalities, which is known as the feature confusion problem of neural networks. The attention of identical object has a good correspondence between RGB and Sonar modalities in Fig. \ref{fig:attention}, proving that the attention relying on RGB object can be transmitted to the corresponding Sonar object, and the CAFusion module effectively implements RGB-Sonar spatial misalignment feature fusion.

\subsubsection{Comparisons on Classes}

We compare the performance of the models on different classes in Table \ref{tab:com_cls}. Our RSFusionDet achieves the best performance on all classes and significantly improves RGB and Sonar detection performance on reflector (+0.2/+0.5 AP) and frustum (+0.7/+0.9 AP) objects, compared to the other best performance. It indicates the effectiveness of RSFusionDet for objects of different shapes.

\subsubsection{Comparisons on Light and Dark Scenes}

We evaluate the performance of the models on the light and dark scenes of the RSFusion validation set, in Table \ref{tab:com_sce}. Our RSFusionDet improves 0.8/1.1 and 1.0/1.2 AP upon DINO~\cite{dino} in light and dark scenes, respectively. Notably, the improvement in RGB detection is greater in dark scenes than in light scenes. In dark scenes, the features of the RGB modality are significantly reduced, while the Sonar modality is unaffected by dark scenes and can consistently provide complementary object features. It demonstrates that the superiority of RSFusionDet in underwater dark scenes, which significantly improves the performance of object detection in dark scenes.

\subsubsection{Comparisons on Maximum Sonar Ranges}

The performance of the models on the validation set with different maximum sonar range image pairs is shown in Table \ref{tab:com_sr}. The results present that the improvement of RSFusionDet at different maximum sonar ranges, and as the maximum sonar range increases, the improvement of RGB detection performance also improves, while the improvement of Sonar detection performance decreases, compared to the best model DINO~\cite{dino}. In the maximum sonar range of 5 and 15 meters, the improvement of detection performance of RGB and Sonar is lower because these two data are more numerous and contain a large number of difficult objects (sr5, sr15, and sr10, sr20 can be separated to observe performance changes), such as small objects, objects in dark scenes, and so on. Excluding the influence of this factor, an increase in object distance results in less object feature information in RGB, while Sonar can provide complementary object feature information for RGB, thereby improving RGB detection performance at large maximum sonar ranges. As the object distance increases, the rich object feature information provided to Sonar by RGB is reduced, and the scale of the object in the Sonar image decreases, resulting in reduced Sonar detection performance at large maximum sonar ranges.

\begin{table}[t]
\caption{Comparisons (AP) on different maximum sonar ranges.}
\centering
\resizebox{\linewidth}{!}{

\begin{tabular}{@{}lccccc@{}}
\toprule
\textbf{Model} &
\textbf{Modality} &
\(\textbf{AP}_{sr5}\) &
\(\textbf{AP}_{sr10}\) &
\(\textbf{AP}_{sr15}\) &
\(\textbf{AP}_{sr20}\) \\ \midrule
Faster R-CNN      & R/S & 72.0 / 55.5 & 84.7 / 46.5 & 70.9 / 37.0 & 74.9 / 39.2 \\
YOLOv3            & R/S & 60.8 / 49.2 & 75.5 / 39.7 & 65.0 / 33.5 & 67.1 / 35.0 \\
FCOS              & R/S & 71.4 / 52.0 & 83.0 / 39.9 & 67.9 / 35.3 & 74.4 / 35.2 \\
ATSS              & R/S & 72.1 / 52.1 & 86.2 / 42.3 & 70.9 / 40.3 & 75.9 / 40.4 \\
Sparse R-CNN      & R/S & 73.9 / 52.1 & 81.2 / 41.6 & 66.8 / 37.5 & 70.7 / 37.8 \\
TOOD              & R/S & 63.8 / 53.8 & 76.6 / 44.4 & 60.4 / 40.4 & 64.0 / 42.2 \\
YOLOX-L           & R/S & 70.4 / 51.2 & 81.3 / 44.4 & 63.7 / 38.9 & 73.3 / 38.2 \\
Deformable DETR++ & R/S & 73.7 / 54.8 & 82.7 / 44.8 & 69.2 / 39.7 & 74.2 / 40.1 \\
DAB-DETR          & R/S & 66.0 / 40.3 & 75.2 / 29.1 & 60.9 / 28.3 & 62.3 / 24.0 \\
DINO              & R/S & 76.6 / 55.7 & 86.4 / 49.4 & 73.0 / 43.4 & 76.7 / 45.6 \\ \midrule
DINO+GAFF\dag     & R+S & 76.7 / 56.2 & 87.0 / 50.1 & 73.1 / 43.8 & 77.4 / 46.0 \\
DINO+SCANet\dag   & R+S & 76.8 / 56.8 & 87.7 / 50.7 & 73.2 / 44.2 & 78.1 / 46.4 \\ \midrule
RSFusionDet       & R+S & 
                  \textbf{76.9} / \textbf{57.3} &
                  \textbf{88.4} / \textbf{51.3} &
                  \textbf{73.4} / \textbf{44.6} &
                  \textbf{78.9} / \textbf{46.9} \\ \bottomrule
                  
\end{tabular}

}
\label{tab:com_sr}
\end{table}

\subsection{Ablation Studies}
\label{sec:as}

\begin{table}[t]
\caption{Ablation study for CAFusion and OMHead. The self-attention represents using self-attention instead of cross-attention in CAFusion, removing the feature fusion of RGB and Sonar.}
\centering
\resizebox{0.9\linewidth}{!}{

\begin{tabular}{@{}cccc@{}}
\toprule
\textbf{CAFusion} & \textbf{OMHead} & \textbf{AP} & \(\textbf{F1-Score}_{match}\) \\ \midrule
                  &                    & 75.7 / 47.2   & - \\
\checkmark self-attention &            & 75.7 / 47.3   & - \\
\checkmark        &                    & 76.2 / 48.3   & - \\
                  & \checkmark         & 75.8 / 47.4   & 70.1 \\
\checkmark self-attention & \checkmark & 75.8 / 47.5   & 70.4 \\
\checkmark        & \checkmark         & \textbf{76.4} / \textbf{48.6}   & \textbf{83.4} \\ \bottomrule
\end{tabular}

}
\label{tab:ab_rsfusiondet}
\end{table}

\subsubsection{Ablation Study on RSFusionDet}

Table \ref{tab:ab_rsfusiondet} shows the ablation of CAFusion and OMHead for RSFusionDet. The CAFusion significantly improves RGB and Sonar detection performance by 0.5/1.1 AP, and the OMHead achieves object matching between RGB and Sonar with an 83.4 \(\text{F1-Score}_{match}\). Besides, we change the cross-attention to self-attention in the CAFusion module (\textit{i.e.} removing RGB-Sonar feature fusion), with the same \#Params and GFLOPs, and find that the performance does not change in RGB modality and only improves 0.1 AP in Sonar modality. The results illustrate the effectiveness of CAFusion and OMHead for RGB-Sonar feature fusion and object matching. And it indicates that the gain of CAFusion purely comes from RGB-Sonar feature fusion rather than from the additional parameter counts and computational complexity introduced by CAFusion. Moreover, OMHead has slightly improved detection performance by 0.2/0.3 AP based on CAFusion, demonstrating that object matching contributes to RGB-Sonar feature fusion. OMHead also improves detection performance by 0.1/0.2 AP without CAFusion (or with self-attention CAFusion), but the object matching performance drops to 70.1 (or 70.4) \(\text{F1-Score}_{match}\), indicating the importance of CAFusion in RGB-Sonar feature fusion. The CAFusion not only achieves RGB-Sonar feature fusion and improves the performance of RGB-Sonar multimodal object detection, but also enhances the correlation of RGB-Sonar features in object matching.

\begin{table}[t]
\caption{Ablation study for fusion methods.}
\centering
\resizebox{0.7\linewidth}{!}{

\begin{tabular}{@{}ccc@{}}
\toprule
\textbf{Fusion Method} & \textbf{AP} & \(\textbf{F1-Score}_{match}\) \\ \midrule
Add                      & 76.0 / 47.9 & 78.9 \\
Concat                   & \textbf{76.4} / \textbf{48.6} & \textbf{83.4} \\ \bottomrule
\end{tabular}

}
\label{tab:ab_fm}
\end{table}
\begin{table}[t]
\caption{Ablation study for the structure of CAFusion.}
\centering
\resizebox{0.8\linewidth}{!}{

\begin{tabular}{@{}lc@{}}
\toprule
\textbf{Module} & \textbf{AP} \\ \midrule
CAFusion                         & \textbf{76.2} / \textbf{48.3} \\ \midrule
CAFusion w/o \(\bm{p}_{pe}\)     & 75.9 / 47.8 \\
CAFusion w/o \(\bm{p}_{sle}\)    & 76.1 / 47.9 \\
CAFusion w/o Linear (in CALayer) & 75.9 / 47.7 \\\bottomrule
\end{tabular}

}
\label{tab:ab_cafusion}
\end{table}

\subsubsection{Ablation Study on CAFusion}

To explore the performance of CAFusion, we perform ablation on its settings. As Eq. \ref{eq:concat}, RGB and aligned Sonar features are fused using the concatenate operation, Sonar and aligned RGB features are the same. In Table \ref{tab:ab_fm}, we perform ablation on the fusion method of add and concatenate operations, and find that the add operation is not as effective as the concatenate operation in both RGB-Sonar multimodal object detection and object matching performance. The fusion method of the add operation can lead to the loss of important detailed information. Even if RGB and Sonar are spatially aligned, their features will still have spatial deviations. The add operation can cause interference between inconsistent areas. Compared to the concatenate operation, the add operation has a weaker ability in representing nonlinear relationships and cannot model more complex interactions between features. In Table \ref{tab:ab_cafusion}, we ablate the structure of CAFusion including positional embeddings \(\bm{p}_{pe}\), scale-level embeddings \(\bm{p}_{sle}\), and Linear in CALayer. The results indicate that without these settings, the performance of RSFusionDet decreases. It also proves that \(\bm{p}_{pe}\) and \(\bm{p}_{sle}\) provide spatial position encoding information and scale-level position information for RGB and Sonar features, which contributes to RGB-Sonar spatial misalignment feature fusion, and the Linear in CALayer provides effective feature encoding in RGB-Sonar feature alignment. Although removing the scale-level embeddings \(\bm{p}_{sle}\) has a relatively small impact on the detection performance of the RGB modality (only reducing 0.1 AP), it has a significant effect on the Sonar modality. This is mainly because the objects in the RGB modality tend to be large-scale, and their multi-scale features contain rich object information. Therefore, the RGB modality is not sensitive to \(\bm{p}_{sle}\). Additionally, we evaluate the CAFusion w/o \(\bm{p}_{sle}\) on a partial dataset containing small objects, showing an improvement of 0.8/0.9 AP. It indicates that \(\bm{p}_{sle}\) can bring greater performance gain on RGB modality for small objects, and \(\bm{p}_{sle}\) proves its significance for both RGB and Sonar modalities. In contrast, \(\bm{p}_{pe}\) has a greater impact on the detection performance of both RGB and Sonar modalities, as it plays a key role in helping RSFusionDet learn the spatial position correspondence between RGB and Sonar object features.

\begin{table}[t]
\caption{Ablation study for the number of layers in CAFusion.}
\centering
\resizebox{0.75\linewidth}{!}{

\begin{tabular}{@{}cccc@{}}
\toprule
\textbf{\#layers} & \textbf{\#Params (M)} & \textbf{GFLOPs} & \textbf{AP} \\ \midrule
1                 & 97                    & 241             & 75.9 / 47.5 \\
2                 & 97                    & 246             & 75.9 / 47.9 \\
\textbf{3}        & 98                    & 251             & \textbf{76.2} / \textbf{48.3} \\
4                 & 99                    & 256             & 76.0 / 47.8 \\
5                 & 99                    & 261             & 76.1 / 48.0 \\
6                 & 100                   & 266             & 76.0 / 47.9 \\ \bottomrule
\end{tabular}

}
\label{tab:ab_calayer}
\end{table}
\begin{table}[t]
\caption{Ablation study for the number of layers (independently in RGB and Sonar feature space) in CAFusion. Vertical and horizontal headers represent the number of layers (\#layers) of CALayer in RGB and Sonar feature spaces, respectively. The metrics are object detection AP in RGB/Sonar modality.}
\centering
\resizebox{\linewidth}{!}{

\begin{tabular}{@{}ccccccc@{}}
\toprule
\textbf{RGB\textbackslash Sonar} &
\textbf{1} &
\textbf{2} &
\textbf{3} &
\textbf{4} &
\textbf{5} &
\textbf{6} \\ \midrule
\textbf{1} & 75.9/47.5 &75.9/48.0 & 75.8/48.3 & 75.8/47.9 & 75.8/47.8 & 75.8/47.8 \\
\textbf{2} & 75.9/47.5 &75.9/47.9 & 75.9/48.3 & 75.9/47.9 & 75.8/47.9 & 75.8/47.9 \\
\textbf{3} & 76.2/47.5 &76.2/47.9 & \textbf{76.2/48.3} & 76.2/47.8 & 76.1/47.7 & 76.1/47.8 \\
\textbf{4} & 76.1/47.4 &76.1/47.9 & 76.1/48.3 & 76.0/47.8 & 76.0/47.9 & 76.0/48.0 \\
\textbf{5} & 76.0/47.5 &76.1/47.8 & 76.1/48.3 & 76.1/47.8 & 76.1/48.0 & 76.0/47.9 \\
\textbf{6} & 76.0/47.3 &76.0/47.8 & 76.0/48.2 & 76.0/47.7 & 76.0/47.8 & 76.0/47.9 \\ \bottomrule
\end{tabular}

}
\label{tab:ab_calayer_rgbs_independent}
\end{table}
\begin{table}[t]
\caption{Ablation study for the number of heads in CAFusion.}
\centering
\resizebox{0.8\linewidth}{!}{

\begin{tabular}{@{}cccc@{}}
\toprule
\textbf{\#heads} & \textbf{\#Params (M)} & \textbf{GFLOPs} & \textbf{AP} \\ \midrule
1                & 97                    & 247             & 76.1 / 48.1 \\
2                & 97                    & 247             & 76.1 / 48.2 \\
\textbf{4}       & 98                    & 249             & \textbf{76.4} / \textbf{48.6} \\
8                & 98                    & 251             & 76.2 / 48.3 \\ \bottomrule
\end{tabular}

}
\label{tab:ab_cahead}
\end{table}

In Table \ref{tab:ab_calayer}, we ablate the cross-attention layers, each of which has only about 5 GFLOPs. The results illustrate that RSFusionDet performs best when the number of layers is 3 in both RGB and Sonar modalities, and that the number of layers has different effects on RGB and Sonar modalities. And we set different layers for RGB and Sonar modalities to further explore the CAFusion module in Table \ref{tab:ab_calayer_rgbs_independent}. It indicates that the number of layers in CAFusion has a relatively small impact on the RGB modality and a great impact on the Sonar modality. This is mainly because the object feature information in the RGB modality is much richer than that in the Sonar modality. Following RGB-Sonar feature fusion, Sonar modality provides a small amount of object feature information for RGB modality, resulting in RGB modality gaining less from RGB-Sonar feature fusion, while Sonar modality gains a large amount. Moreover, the imbalance of layers in CAFusion between RGB and Sonar modalities can slightly reduce model performance, primarily due to the learning imbalance between the RGB and Sonar modalities. The results show that RSFusionDet performs best when the number of layers in the RGB and Sonar modalities of CAFusion is balanced at 3. In CAFusion, RGB-Sonar features have highly unaligned features, and the number of heads in the cross-attention split feature vectors. We believe that the number of heads may affect the correlation between features, and thus affect the RGB-Sonar feature fusion. In Table \ref{tab:ab_cahead}, we reduce the number of heads for attention to increase the receptive field of the RGB-Sonar feature correlation calculation. The results show that when the number of heads is 4, RSFusionDet has the best performance and slightly reduces the computational complexity of attention.

\begin{table}[t]
\caption{Ablation study for object matching strategies.}
\centering
\resizebox{\linewidth}{!}{

\begin{tabular}{@{}ccccc@{}}
\toprule
\textbf{BBox priori} &
\textbf{IoU weight} &
\(\textbf{P}_{match}\) &
\(\textbf{R}_{match}\) &
\(\textbf{F1-Score}_{match}\) \\ \midrule
           &            & 8.6  & 8.1  & 8.4  \\
\checkmark &            & 17.2 & 16.7 & 16.9 \\
           & \checkmark & 77.2 & 83.8 & 80.4 \\
\checkmark & \checkmark & \textbf{80.9} & \textbf{86.0} & \textbf{83.4} \\ \bottomrule
\end{tabular}

}
\label{tab:ab_omhs}
\end{table}
\begin{table}[t]
\caption{Ablation study for the number of layers in OMHead.}
\centering
\resizebox{\linewidth}{!}{

\begin{tabular}{@{}ccccc@{}}
\toprule
\textbf{\#layers} &
\textbf{\#Params (M)} &
\textbf{GFLOPs} &
\textbf{AP} &
\(\textbf{F1-Score}_{match}\) \\ \midrule
1          & 98 & 251 & 75.9 / 47.8 & 80.3 \\
2          & 98 & 251 & 76.1 / 47.9 & 80.7 \\
\textbf{3} & 98 & 251 & \textbf{76.2} / \textbf{48.3} & \textbf{83.4} \\
4          & 98 & 251 & 76.2 / 48.2 & 79.7 \\
5          & 98 & 251 & 76.1 / 47.9 & 80.2 \\
6          & 98 & 252 & 75.8 / 47.9 & 79.9 \\ \bottomrule
\end{tabular}

}
\label{tab:ab_omhlayer}
\end{table}
\begin{table}[t]
\caption{Ablation study for the filtering and matching threshold in OMHead. Vertical and horizontal headers represent the filtering and matching threshold, respectively. The metric is \(\text{F1-Score}_{match}\).}
\centering
\resizebox{\linewidth}{!}{

\begin{tabular}{@{}cccccccccc@{}}
\toprule
\textbf{filter\textbackslash match} &
\textbf{0.1} &
\textbf{0.2} &
\textbf{0.3} &
\textbf{0.4} &
\textbf{0.5} &
\textbf{0.6} &
\textbf{0.7} &
\textbf{0.8} &
\textbf{0.9} \\ \midrule
\textbf{0.1} & 80.0 & 80.3 & 80.5 & 80.7 & 81.0 & 81.0 & 81.2 & 81.5 & 81.5 \\
\textbf{0.2} & 81.6 & 82.1 & 82.3 & 82.5 & 82.7 & 82.7 & 82.9 & 82.9 & 82.8 \\
\textbf{0.3} & 82.2 & 82.7 & 82.6 & 83.0 & 83.2 & 83.2 & 83.3 & \textbf{83.4} & 83.2 \\
\textbf{0.4} & 82.2 & 82.6 & 82.7 & 82.9 & 83.0 & 83.1 & 83.3 & 83.3 & 83.1 \\
\textbf{0.5} & 82.2 & 82.5 & 82.6 & 82.8 & 83.0 & 83.1 & 83.3 & 83.2 & 83.0 \\
\textbf{0.6} & 81.1 & 81.4 & 81.7 & 81.9 & 81.9 & 82.0 & 82.2 & 82.3 & 82.3 \\
\textbf{0.7} & 80.1 & 80.4 & 80.8 & 81.0 & 81.0 & 80.9 & 81.1 & 81.3 & 81.3 \\
\textbf{0.8} & 78.1 & 78.4 & 78.8 & 79.0 & 79.0 & 79.1 & 79.2 & 79.5 & 79.4 \\
\textbf{0.9} & 59.7 & 60.0 & 60.3 & 60.4 & 60.4 & 60.5 & 60.5 & 60.6 & 60.6 \\ \bottomrule
\end{tabular}

}
\label{tab:ab_ofm_threshold}
\end{table}

\subsubsection{Ablation Study on OMHead}

To demonstrate the effectiveness of the RGB-Sonar object matching method, we perform ablation on object matching strategies in Table \ref{tab:ab_omhs}. The results indicate that adding bounding box priori information to the object feature slightly improves the matching performance of 8.5 \(\text{F1-Score}_{match}\), and optimizing the object matching loss using IoU weight can significantly improve the matching performance of 72 \(\text{F1-Score}_{match}\). Finally, with the optimization of these two strategies, object matching achieves the best performance of 83.4 \(\text{F1-Score}_{match}\). The bounding box priori information provides object position information for RGB and Sonar modalities, which is beneficial for OMHead modelling RGB-Sonar object position mapping associations, thereby improving object matching performance. Furthermore, the IoU weight adaptively assigns learning weights to object matching features based on the richness of the object feature information, which reduces the impact of noise during the OMHead training process and improves the learning efficiency of the OMHead, thereby significantly improving its object matching performance. Meanwhile, we perform ablation on the encoding layers (MLP) of the OMHead object features in Table \ref{tab:ab_omhlayer} and find that the best object detection and object matching performance is achieved when the number of encoding layers is 3. The results also indicate that object matching can slightly affect the effect of RGB-Sonar feature fusion, while OMHead has a tiny number of parameters and computational complexity. Besides, we perform ablation on the filtering and matching threshold in OMHead, as shown in Table \ref{tab:ab_ofm_threshold}. The results illustrate that the object matching performance of OMHead is mainly affected by the filtering threshold. When the filtering threshold is too low (\textless 0.3), a large number of low-confidence erroneous objects are introduced to OMHead. Conversely, when the filtering threshold is too high (\textgreater 0.5), some low-confidence correct objects are filtered out, leading to a decrease in the object matching performance of OMHead. In contrast, the matching threshold has a relatively small impact on the object matching performance of OMHead. As the matching threshold increases from 0.1 to 0.8, the object matching performance of OMHead gradually improves. When the matching threshold reaches 0.9, due to its high requirements for matching scores, correct matching results with low matching scores are filtered out, thereby reducing the object matching performance of OMHead. We find that when the filtering threshold is 0.3 and the matching threshold is 0.8, OMHead performs best with 80.9 \(\text{P}_{match}\), 86.0 \(\text{R}_{match}\), 83.4 \(\text{F1-Score}_{match}\).
\section{Conclusion}
\label{sec:conclusion}

In this paper, we put forward an underwater RGB-Sonar multimodal object detection dataset called RSFusion and its evaluation metrics, which is a new benchmark for RGB-Sonar multimodal object detection. Meanwhile, we propose a new RGB-Sonar multimodal object detection architecture called RSFusionDet with a new RGB-Sonar multimodal object detection result expression. The CAFusion and OMHead with OMLoss we design effectively achieve RGB-Sonar spatial misalignment feature fusion (by deformable attention) and object matching. Our results have shown the effectiveness and better performance of RSFusionDet on RGB-Sonar multimodal object detection. The experimental results of our RSFusion benchmark also demonstrate that RGB-Sonar multimodal object detection has good adaptability and high performance improvement for objects in dark scenes, small and medium objects, and distant objects.

Although our RSFusionDet has achieved good performance and improvement on the RSFusion benchmark, further exploration and research are still needed for RGB-Sonar multimodal object detection. We have conducted pioneering research on RGB-Sonar multimodal object detection and provided an RSFusion benchmark. We look forward to further work in the future that can provide better ideas and methods on this topic.

\medskip
\noindent
\textbf{Acknowledgments.}
This research is funded by the National Natural Science Foundation of China, grant number 52371350, and by the National Key Research and Development Program of China, grant number 2023YFC2809104.
{
    \small
    \bibliographystyle{ieeenat_fullname}
    \bibliography{main}
}

\clearpage
\setcounter{page}{1}
\maketitlesupplementary

\section{Details of RSFusion Dataset}
\label{sec:drd}

We provide a detailed introduction and analysis of the RSFusion dataset, including visualization of all objects in RGB and Sonar modalities, processing of sonar raw images, analysis of object properties and distributions, and explanation of annotations.

\subsection{Visualization of All Objects}
\label{sec:vao}

\begin{figure}[t]
\centering

\begin{subfigure}{0.47\linewidth}
\centering
\includegraphics[width=1.0\linewidth]{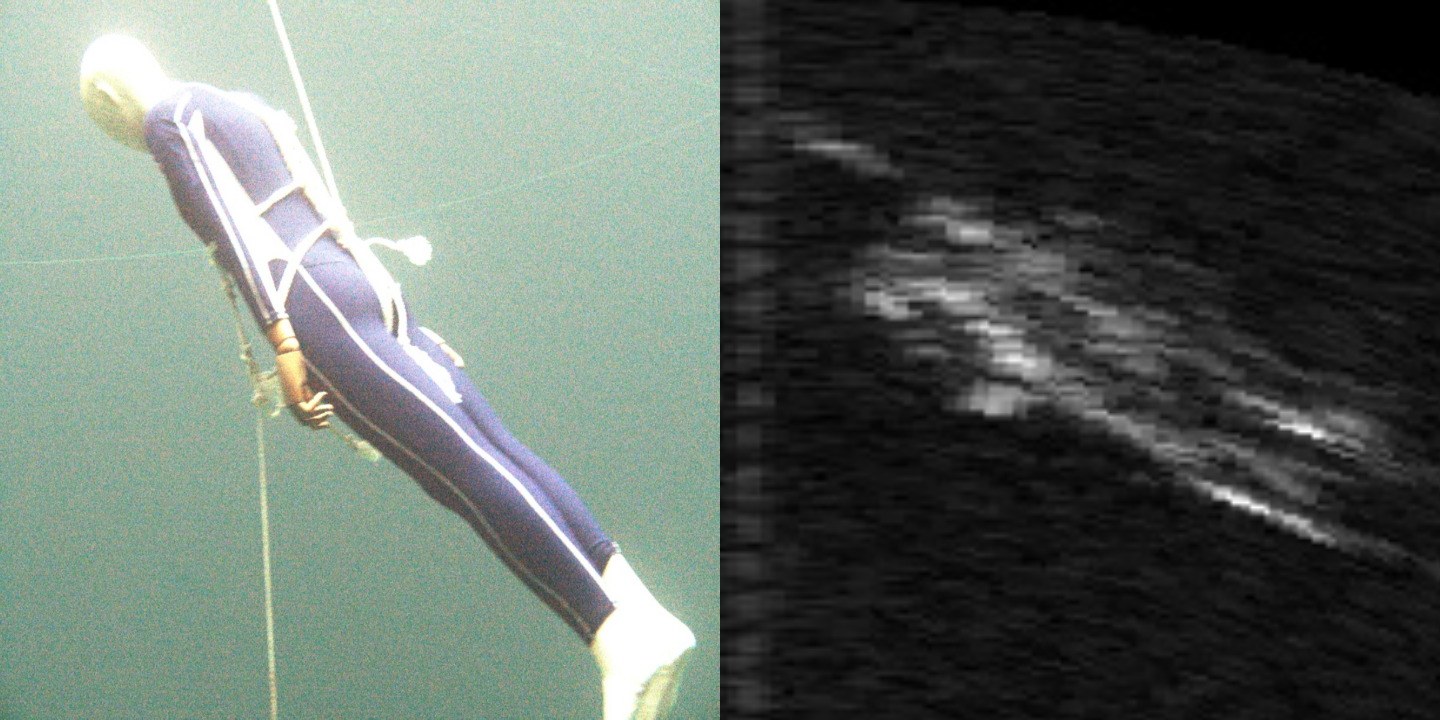}
\caption{mannequin.}
\end{subfigure}
\begin{subfigure}{0.47\linewidth}
\centering
\includegraphics[width=1.0\linewidth]{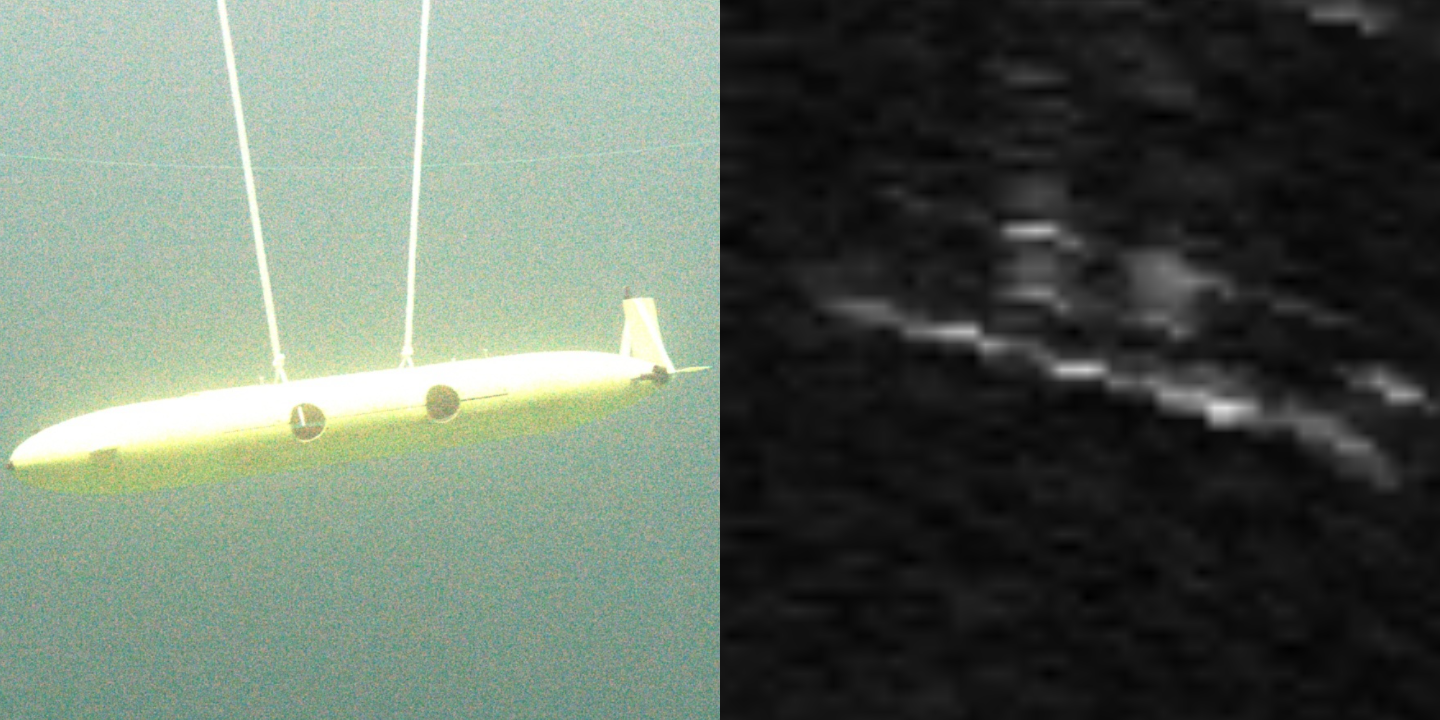}
\caption{UUV.}
\end{subfigure}

\vspace{0pt}

\begin{subfigure}{0.47\linewidth}
\centering
\includegraphics[width=1.0\linewidth]{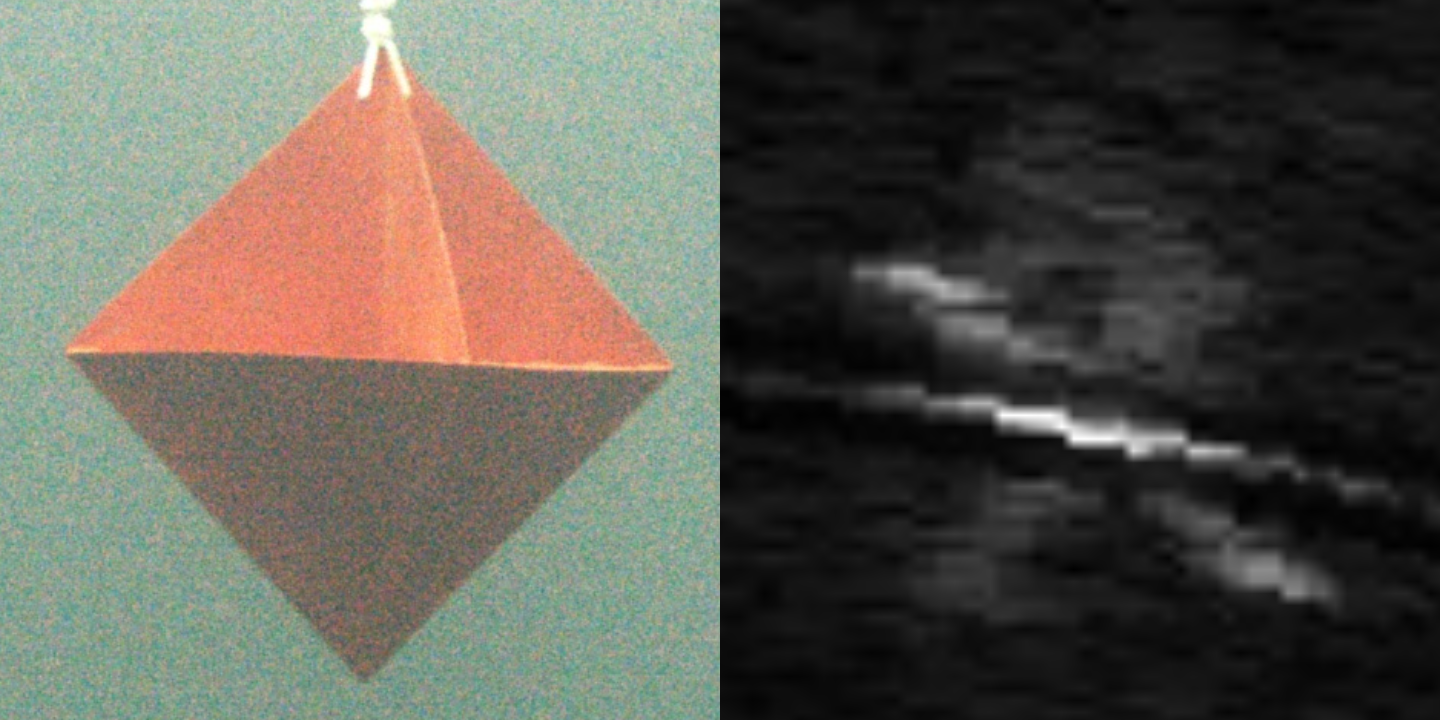}
\caption{reflector.}
\end{subfigure}
\begin{subfigure}{0.47\linewidth}
\centering
\includegraphics[width=1.0\linewidth]{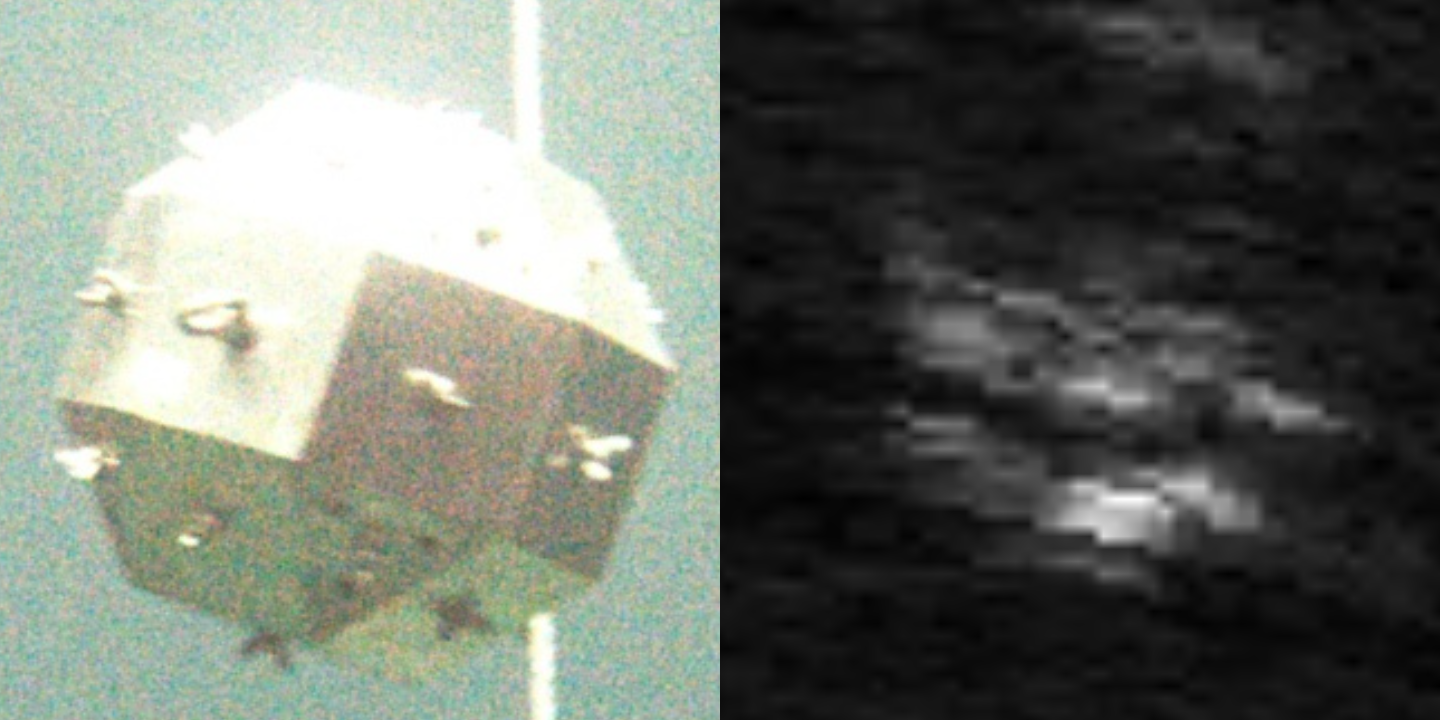}
\caption{metal polyhedron.}
\end{subfigure}

\vspace{0pt}

\begin{subfigure}{0.47\linewidth}
\centering
\includegraphics[width=1.0\linewidth]{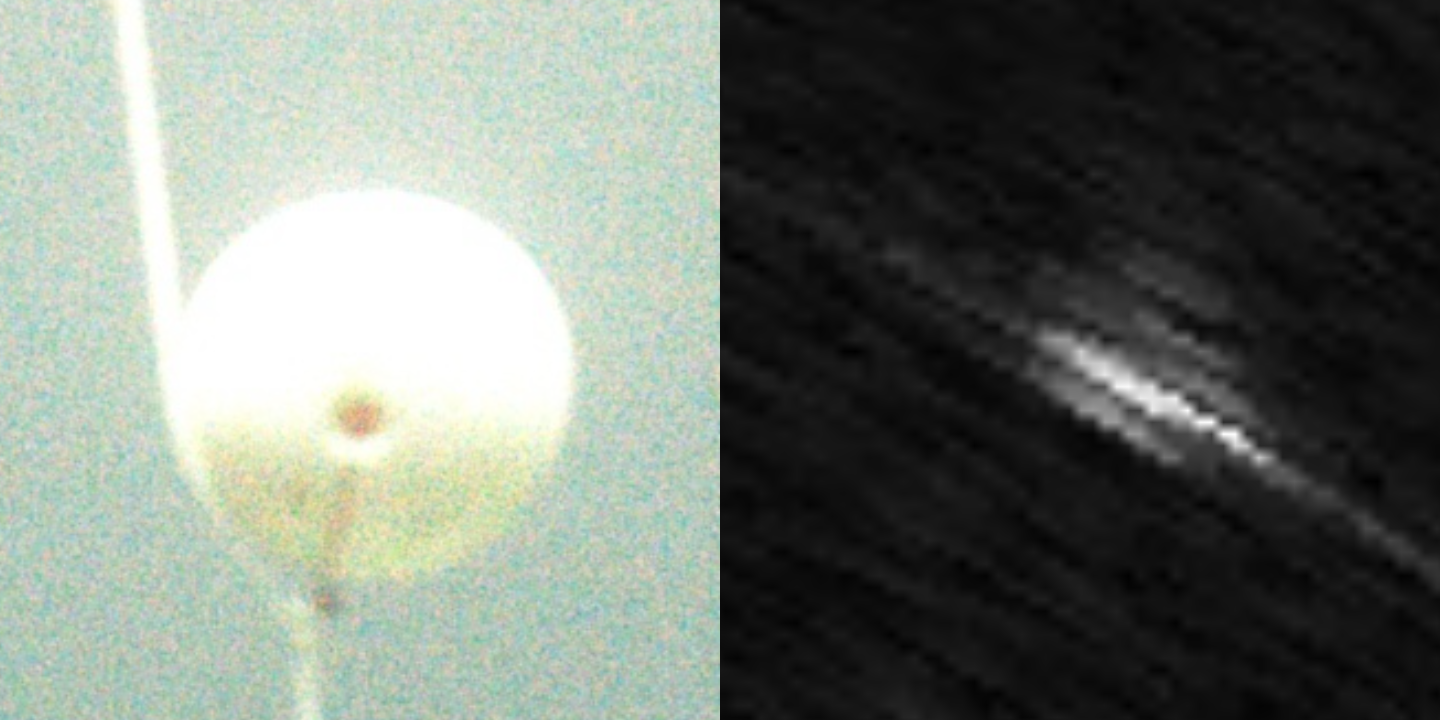}
\caption{float.}
\end{subfigure}
\begin{subfigure}{0.47\linewidth}
\centering
\includegraphics[width=1.0\linewidth]{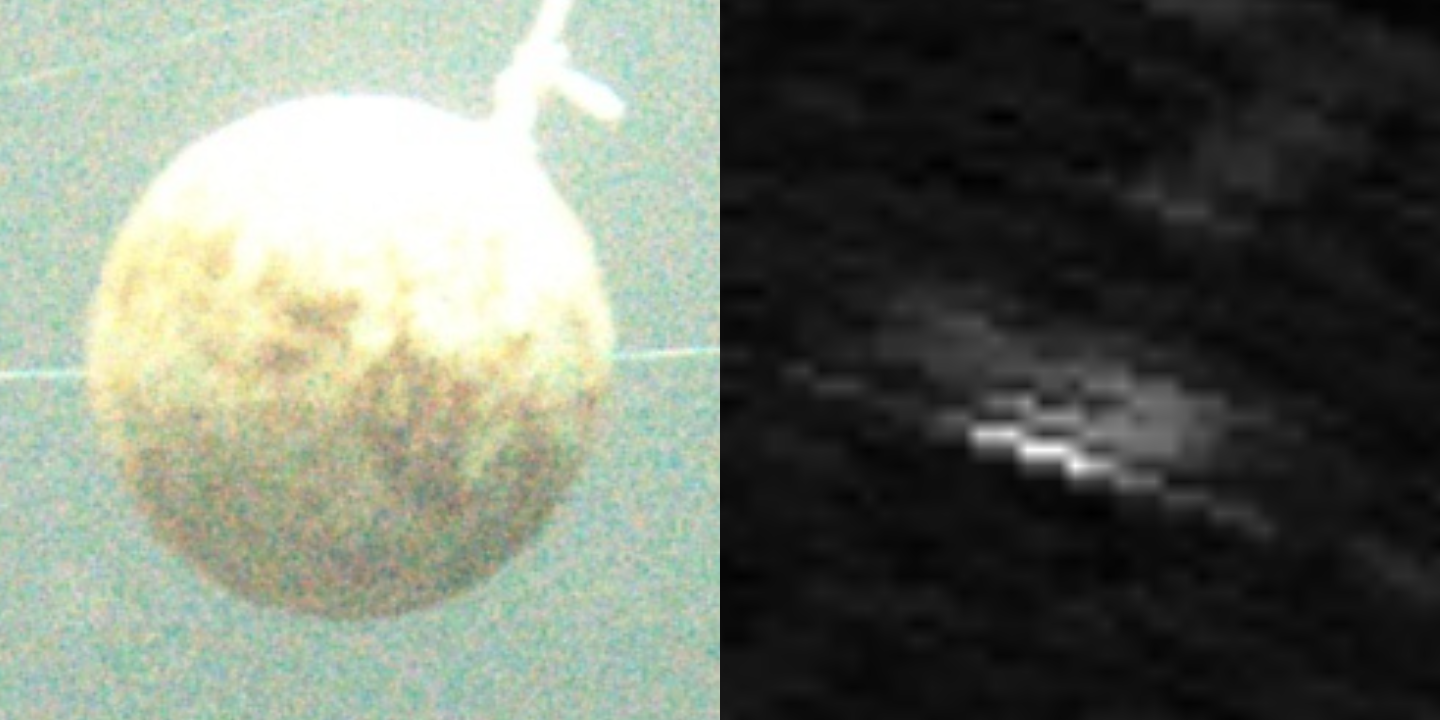}
\caption{iron ball.}
\end{subfigure}

\vspace{0pt}

\begin{subfigure}{0.47\linewidth}
\centering
\includegraphics[width=1.0\linewidth]{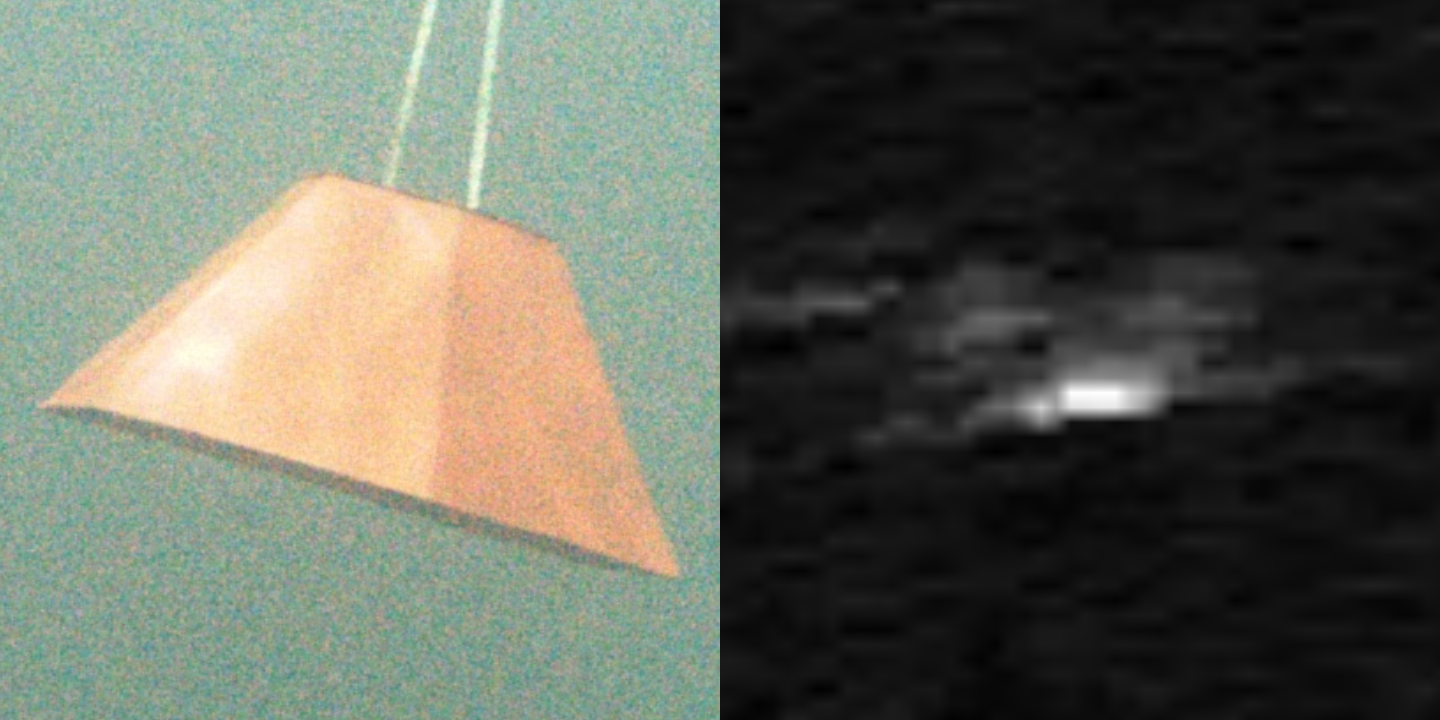}
\caption{frustum.}
\end{subfigure}

\caption{All object classes in the RSFusion dataset. The object on the left is in RGB, while the one on the right is in Sonar.}
\label{fig:RSFusion_objs}
\end{figure}

We visualize all object classes with RGB and Sonar modalities in the RSFusion dataset, in Fig. \ref{fig:RSFusion_objs}. Objects in RGB modality have richer features than those in Sonar modality. It is obvious that there is a significant difference in the object features between the RGB and Sonar modalities, and the features have the spatial feature misalignment problem.

\subsection{Processing of Sonar Raw Images}
\label{sec:psri}

\begin{figure*}[t]
\centering
\includegraphics[width=0.84\linewidth]{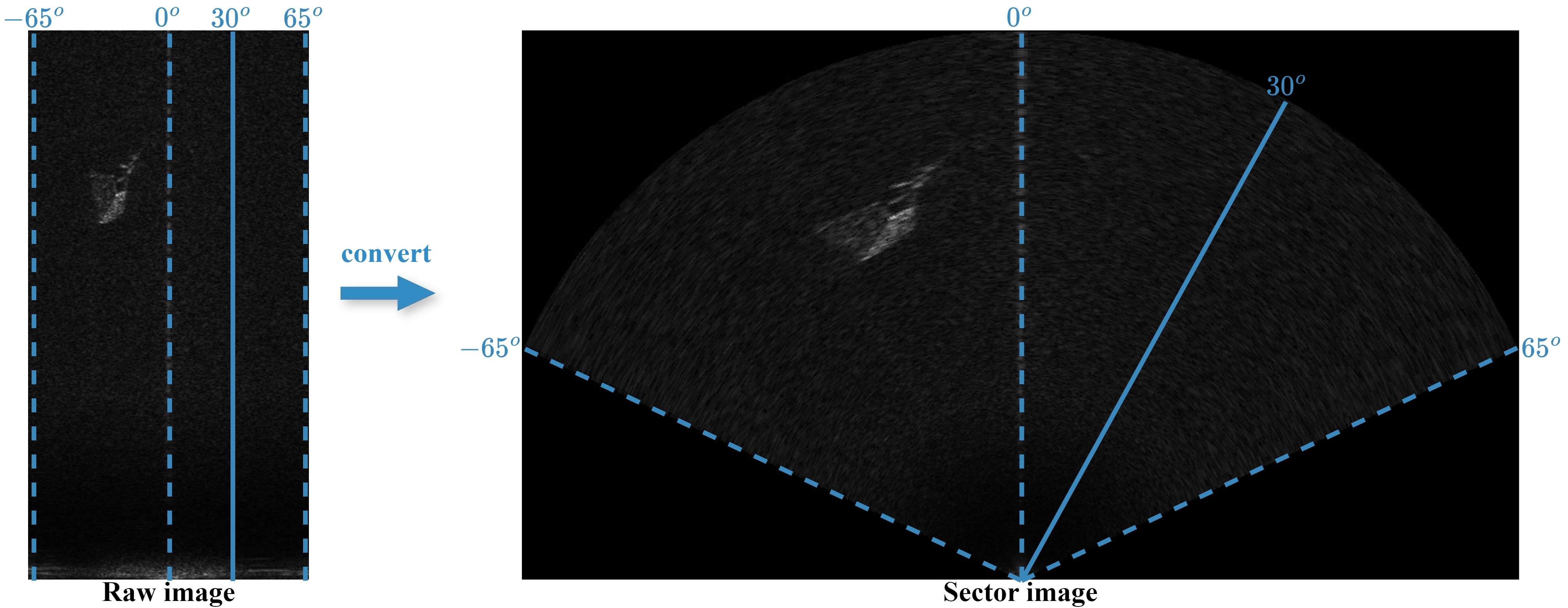}
\caption{Conversion of sonar images from raw to sector.}
\label{fig:csonar}
\end{figure*}

The sonar image includes raw and sector images in the RSFusion dataset, where the sector image is converted from the raw image. The brief conversion process is shown in Fig. \ref{fig:csonar}. The sonar sensor returns a raw image and a list of angles when collecting sonar images. The angle is the horizontal angle of the beam in the forward-looking multibeam sonar (the horizontal field of view of the sonar is 130 degrees), and the echo intensity obtained by sending each beam forward at different vertical angles forms a column of pixel values in the height direction in the raw image. Expand each column of pixels according to their corresponding angles and use nearest neighbour interpolation to fill in missing angles to obtain the converted sonar sector image.


\subsection{Analysis of Object Properties and Distributions}
\label{sec:aopd}

\begin{figure}[t]
\centering

\begin{subfigure}{1.0\linewidth}
\centering
\includegraphics[width=1.0\linewidth]{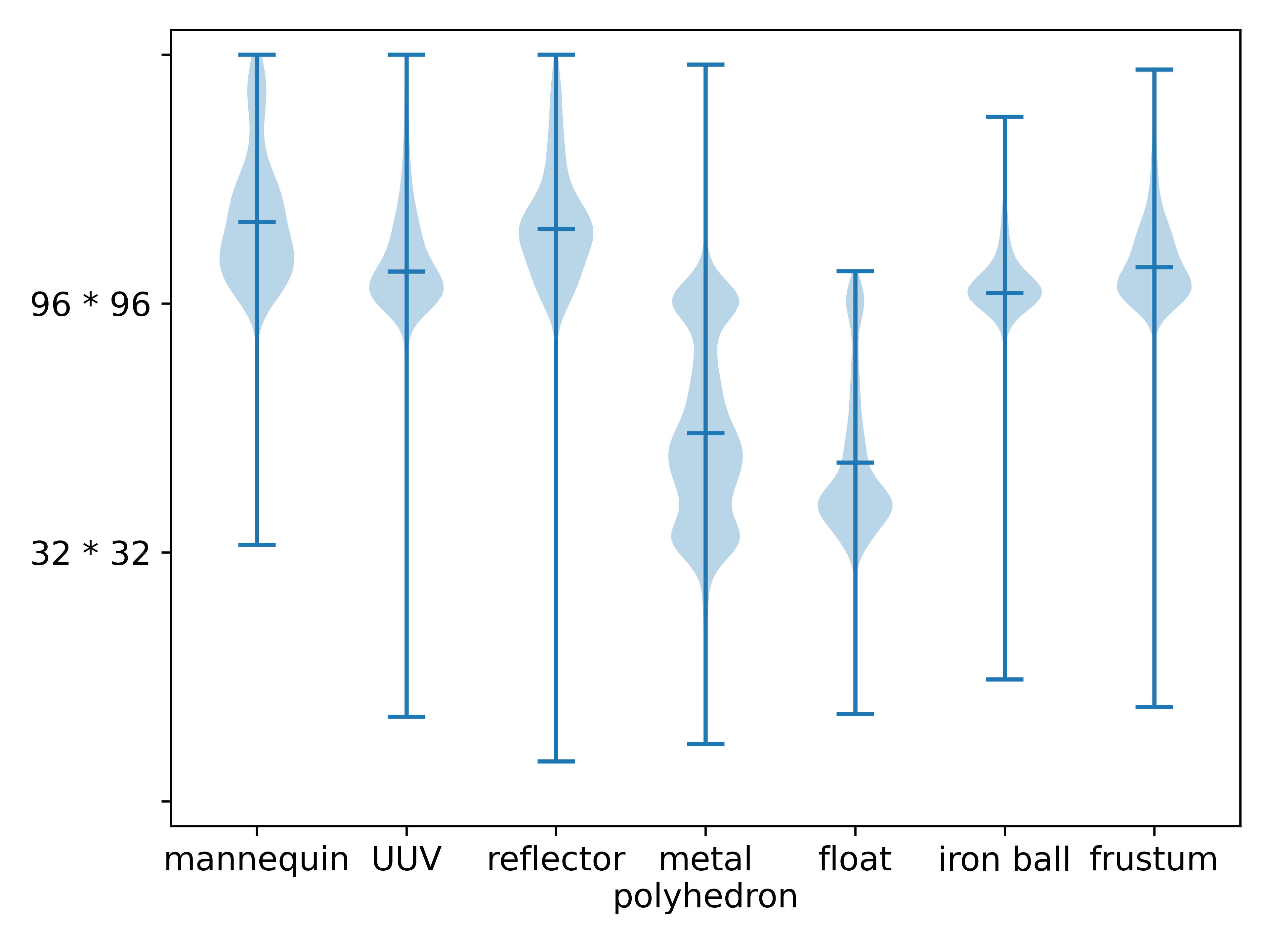}
\caption{Object scales in RGB modality.}
\end{subfigure}

\vspace{0pt}

\begin{subfigure}{1.0\linewidth}
\centering
\includegraphics[width=1.0\linewidth]{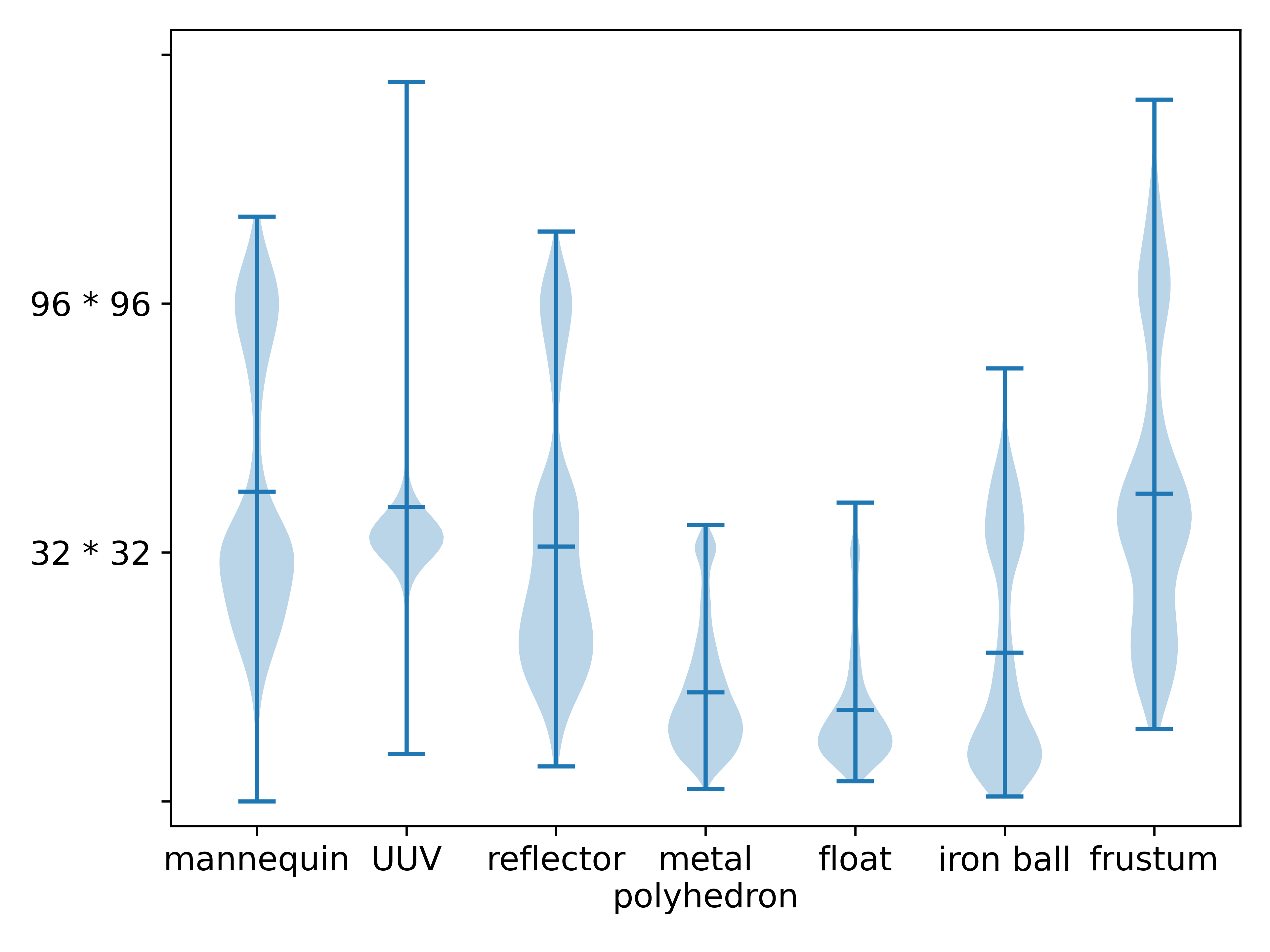}
\caption{Object scales in Sonar modality.}
\end{subfigure}

\caption{Distribution of object scales in the RSFusion dataset.}
\label{fig:obj_scale_ana}
\end{figure}

\noindent
\textbf{Object Scale Analysis.}
We analyze the object scale in our RSFusion dataset based on the object scale specifications of the COCO~\cite{coco} dataset. As shown in Fig. \ref{fig:obj_scale_ana}, the object pixel area is defined as small objects if it is less than 32 * 32, medium objects if it is greater than 32 * 32 and less than 96 * 96, and large objects if it is greater than 96 * 96. In RGB modularity, objects are predominantly large, with metal polyhedron and float concentrated in medium objects, and the distribution of metal polyhedron is more uniform. In the sonar modality, objects are predominantly small and medium.

\begin{figure}[t]
\centering

\begin{subfigure}{1.0\linewidth}
\centering
\includegraphics[width=1.0\linewidth]{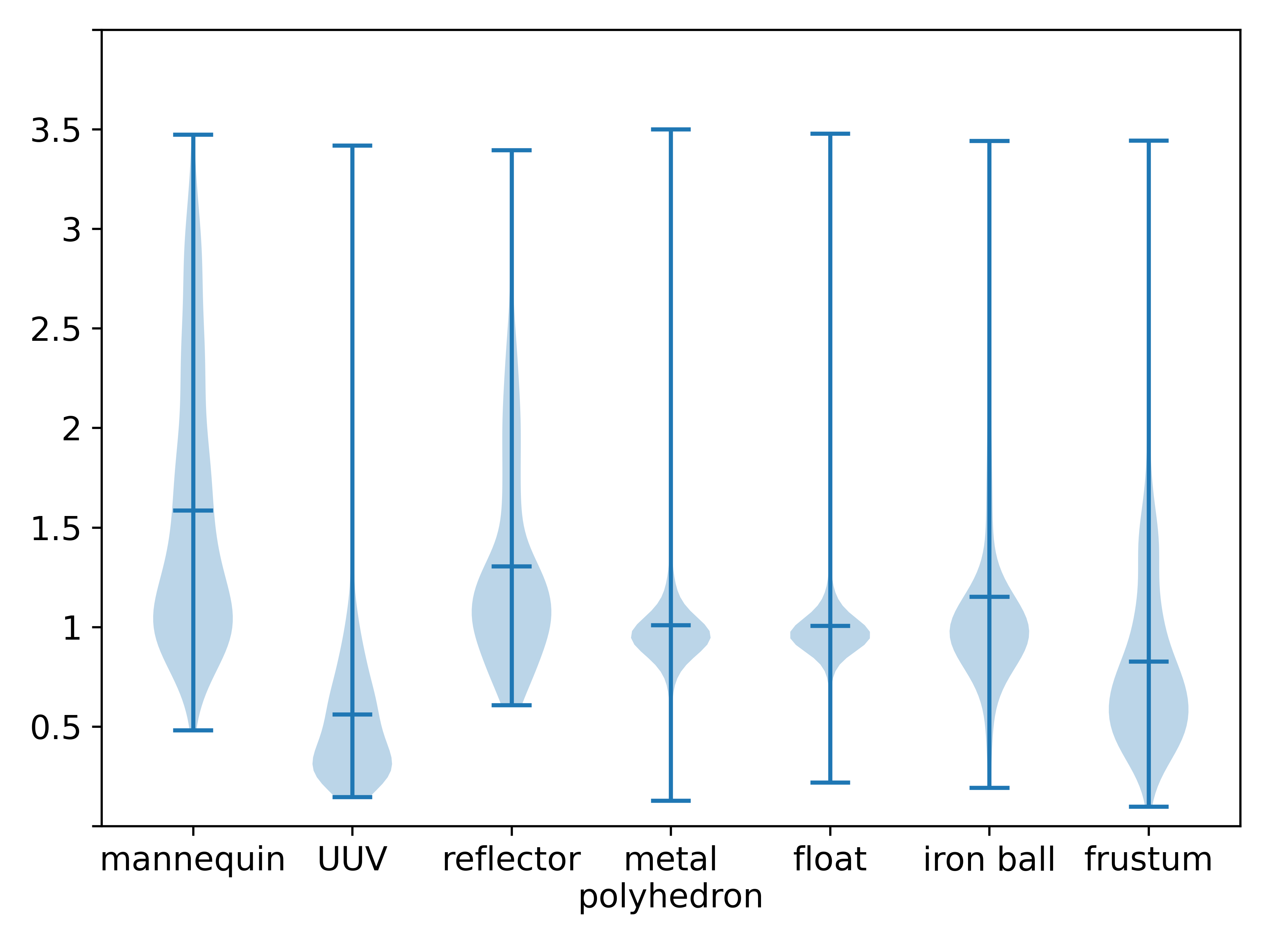}
\caption{Object aspect ratios in RGB modality.}
\end{subfigure}

\vspace{0pt}

\begin{subfigure}{1.0\linewidth}
\centering
\includegraphics[width=1.0\linewidth]{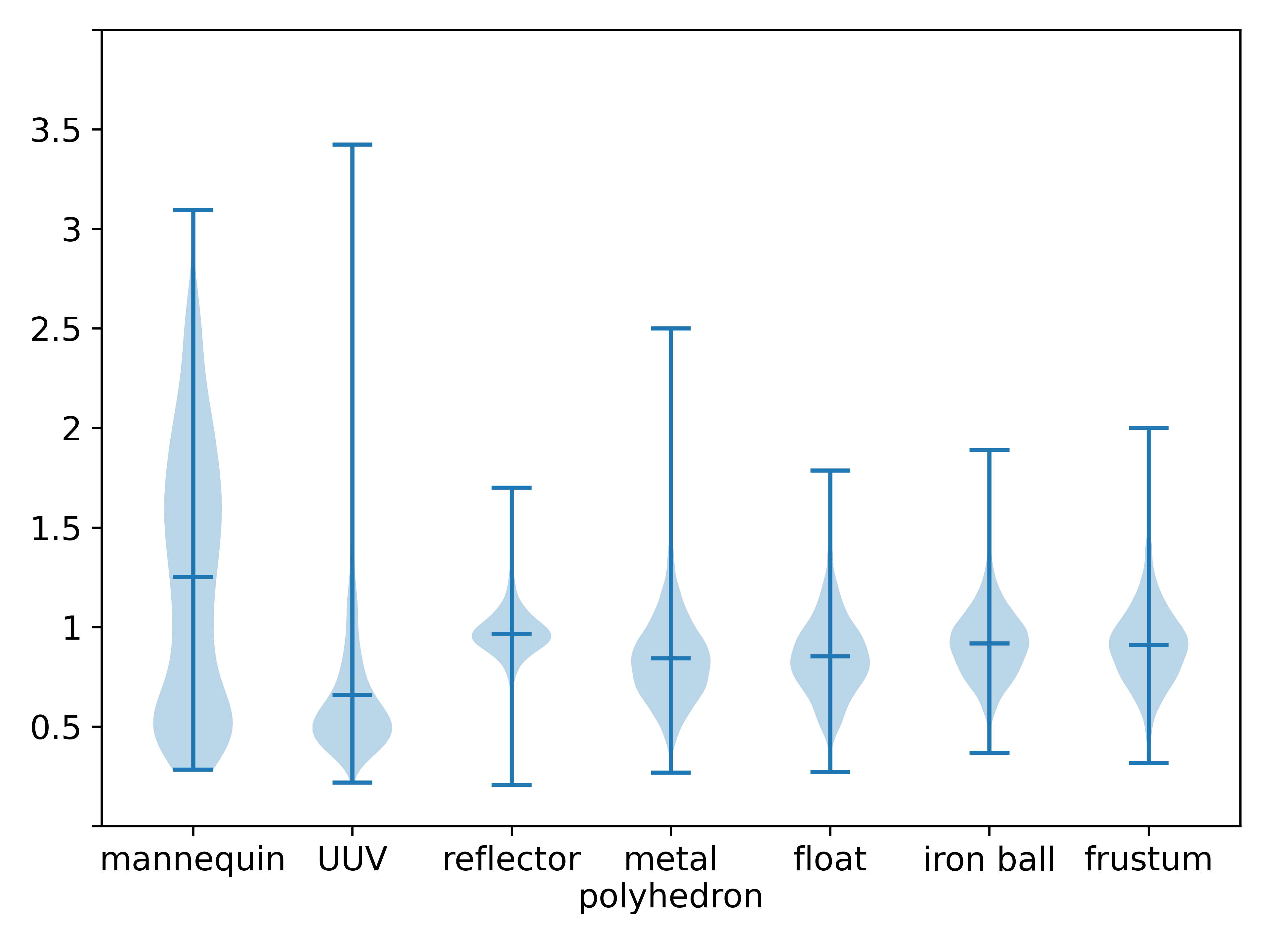}
\caption{Object aspect ratios in Sonar modality.}
\end{subfigure}

\caption{Distribution of object aspect ratios in the RSFusion dataset. The aspect ratio is calculated from \(h / w\) of the object.}
\label{fig:obj_aspect_ratio_ana}
\end{figure}

\noindent
\textbf{Object Aspect Ratio Analysis.}
In Fig. \ref{fig:obj_aspect_ratio_ana}, the object aspect ratios in the RGB and Sonar modalities have a generally consistent distribution, with aspect ratios mainly distributed around 1. Due to the characteristics of the object shape, the aspect ratios of the mannequin are evenly distributed in the range of 0.5-3.0. In contrast, the UUVs are mainly distributed in the range of smaller aspect ratios.

\begin{figure}[t]
\centering
\includegraphics[width=1.0\linewidth]{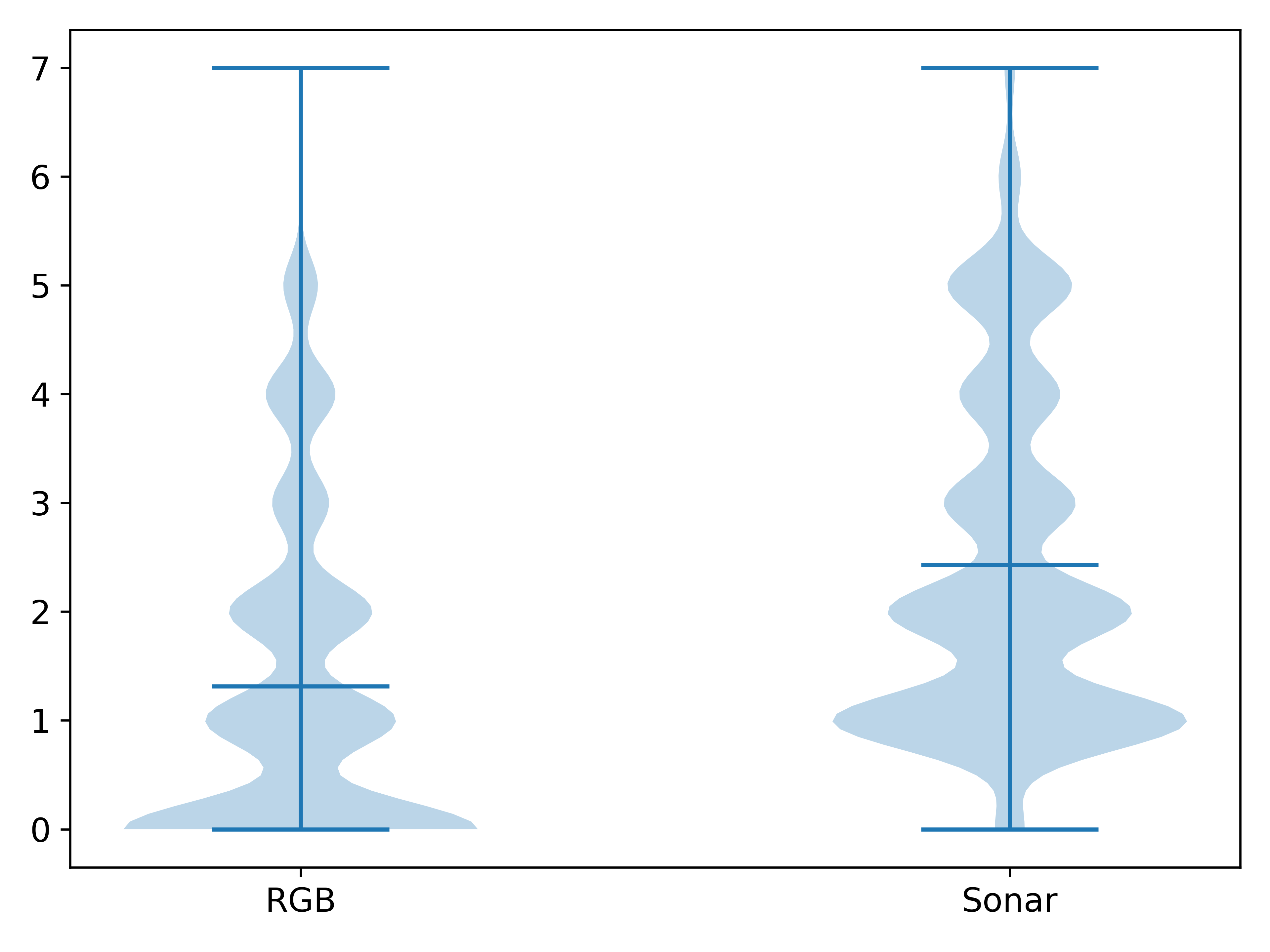}
\caption{Distribution of the number of objects per image in the RSFusion dataset.}
\label{fig:obj_num_per_image_ana}
\end{figure}

\noindent
\textbf{Number of Objects per Image.}
As shown in Fig. \ref{fig:obj_num_per_image_ana}, most RGB images have 0, 1, or 2 objects, while Sonar images have a greater number of objects with an average of 2.5 objects per image, which is 1 more than the average number of objects in RGB images. At the same time, the maximum number of objects per image for both is 7. There are samples in RSFusion where one modality image has no object while the other modality image has objects, but there are no samples where both modalities have no object. Besides, the average number of objects in Sonar images is greater than in RGB images, indicating that some objects visible in Sonar images are invisible in RGB images. It is because the field of view (FOV) of the optical camera is smaller than that of the forward-looking sonar, and sometimes the visible distance of the sonar is greater than that of the optical camera during data collection.

\noindent
\textbf{Distribution of the Number of Objects.}
In Tables \ref{tab:sr_scale}, \ref{tab:sr_scene}, \ref{tab:scene_scale}, and \ref{tab:scene_sr}, we analyze the distribution of the object quantity under different conditions. There are 9298 and 17179 objects in the RGB and Sonar images respectively, and the number of objects in the light scene is about three times that of the dark scene. At the same time, the number of objects in sr15 is the greatest, far greater than in sr5, sr10, and sr20. The srx represents that the maximum sonar range is x meters. There are more small objects in sr5 and sr15, and the number of objects is also greater than in sr10 and sr20, including objects in dark scenes.

\subsection{Explanation of Annotations}
\label{sec:eanns}

Here we present the annotation of our RSFusion dataset. We divide RSFusion equally into training and validation sets in an 8:2 ratio and calculate the mean and std of all the RGB and Sonar images in the dataset, which are mean (132.779, 145.418, 126.863), std (21.792, 20.260, 25.153) for RGB images whose channel order is BGR, and mean 11.451, std 11.565 for Sonar grayscale images. The annotation contains \textit{instances\_train\_images.json}, \textit{instances\_train\_sonars.json}, \textit{instances\_val\_images.json}, and \textit{instances\_val\_sonars.json}. Each annotation has the same structure, which contains two parts: \texttt{metainfo} and \texttt{data\_list}. \texttt{metainfo} includes all kinds of \texttt{classes}, \texttt{environments}, and \texttt{fusion\_flags} information. \texttt{data\_list} contains all the image and object information, as well as the flag information \texttt{fusion\_flag} that indicates whether there is a corresponding identical object in RGB or Sonar modality. In addition, the sonar annotation also includes \texttt{sonar\_info} which contains several sonar metadata. Its details are in our code and dataset.

\begin{table*}[t]
\caption{Distribution of the number of objects in maximum sonar ranges with object scales.}
\centering
\resizebox{\linewidth}{!}{

\begin{tabular}{@{}cccccccccccc|cccccccccccc@{}}
\toprule
\multicolumn{12}{c|}{\textbf{RGB}} &
  \multicolumn{12}{c}{\textbf{Sonar}} \\ \midrule
\multicolumn{3}{c|}{\begin{tabular}[c]{@{}c@{}}sr5\\ 1964\end{tabular}} &
  \multicolumn{3}{c|}{\begin{tabular}[c]{@{}c@{}}sr10\\ 1366\end{tabular}} &
  \multicolumn{3}{c|}{\begin{tabular}[c]{@{}c@{}}sr15\\ 4822\end{tabular}} &
  \multicolumn{3}{c|}{\begin{tabular}[c]{@{}c@{}}sr20\\ 1146\end{tabular}} &
  \multicolumn{3}{c|}{\begin{tabular}[c]{@{}c@{}}sr5\\ 2684\end{tabular}} &
  \multicolumn{3}{c|}{\begin{tabular}[c]{@{}c@{}}sr10\\ 1900\end{tabular}} &
  \multicolumn{3}{c|}{\begin{tabular}[c]{@{}c@{}}sr15\\ 10552\end{tabular}} &
  \multicolumn{3}{c}{\begin{tabular}[c]{@{}c@{}}sr20\\ 2043\end{tabular}} \\ \midrule
\multicolumn{1}{c|}{\begin{tabular}[c]{@{}c@{}}small\\ 170\end{tabular}} &
  \multicolumn{1}{c|}{\begin{tabular}[c]{@{}c@{}}medium\\ 1023\end{tabular}} &
  \multicolumn{1}{c|}{\begin{tabular}[c]{@{}c@{}}large\\ 771\end{tabular}} &
  \multicolumn{1}{c|}{\begin{tabular}[c]{@{}c@{}}small\\ 0\end{tabular}} &
  \multicolumn{1}{c|}{\begin{tabular}[c]{@{}c@{}}medium\\ 806\end{tabular}} &
  \multicolumn{1}{c|}{\begin{tabular}[c]{@{}c@{}}large\\ 560\end{tabular}} &
  \multicolumn{1}{c|}{\begin{tabular}[c]{@{}c@{}}small\\ 27\end{tabular}} &
  \multicolumn{1}{c|}{\begin{tabular}[c]{@{}c@{}}medium\\ 2345\end{tabular}} &
  \multicolumn{1}{c|}{\begin{tabular}[c]{@{}c@{}}large\\ 2450\end{tabular}} &
  \multicolumn{1}{c|}{\begin{tabular}[c]{@{}c@{}}small\\ 0\end{tabular}} &
  \multicolumn{1}{c|}{\begin{tabular}[c]{@{}c@{}}medium\\ 822\end{tabular}} &
  \begin{tabular}[c]{@{}c@{}}large\\ 324\end{tabular} &
  \multicolumn{1}{c|}{\begin{tabular}[c]{@{}c@{}}small\\ 302\end{tabular}} &
  \multicolumn{1}{c|}{\begin{tabular}[c]{@{}c@{}}medium\\ 1455\end{tabular}} &
  \multicolumn{1}{c|}{\begin{tabular}[c]{@{}c@{}}large\\ 927\end{tabular}} &
  \multicolumn{1}{c|}{\begin{tabular}[c]{@{}c@{}}small\\ 1029\end{tabular}} &
  \multicolumn{1}{c|}{\begin{tabular}[c]{@{}c@{}}medium\\ 870\end{tabular}} &
  \multicolumn{1}{c|}{\begin{tabular}[c]{@{}c@{}}large\\ 1\end{tabular}} &
  \multicolumn{1}{c|}{\begin{tabular}[c]{@{}c@{}}small\\ 7944\end{tabular}} &
  \multicolumn{1}{c|}{\begin{tabular}[c]{@{}c@{}}medium\\ 2608\end{tabular}} &
  \multicolumn{1}{c|}{\begin{tabular}[c]{@{}c@{}}large\\ 0\end{tabular}} &
  \multicolumn{1}{c|}{\begin{tabular}[c]{@{}c@{}}small\\ 1479\end{tabular}} &
  \multicolumn{1}{c|}{\begin{tabular}[c]{@{}c@{}}medium\\ 564\end{tabular}} &
  \begin{tabular}[c]{@{}c@{}}large\\ 0\end{tabular} \\ \bottomrule
\end{tabular}

}
\label{tab:sr_scale}
\end{table*}
\begin{table}[t]
\caption{Distribution of the number of objects in maximum sonar ranges with scenes.}
\centering
\resizebox{\linewidth}{!}{

\begin{tabular}{@{}cccccccc|cccccccc@{}}
\toprule
\multicolumn{8}{c|}{\textbf{RGB}} &
  \multicolumn{8}{c}{\textbf{Sonar}} \\ \midrule
\multicolumn{2}{c|}{\begin{tabular}[c]{@{}c@{}}sr5\\ 1964\end{tabular}} &
  \multicolumn{2}{c|}{\begin{tabular}[c]{@{}c@{}}sr10\\ 1366\end{tabular}} &
  \multicolumn{2}{c|}{\begin{tabular}[c]{@{}c@{}}sr15\\ 4822\end{tabular}} &
  \multicolumn{2}{c|}{\begin{tabular}[c]{@{}c@{}}sr20\\ 1146\end{tabular}} &
  \multicolumn{2}{c|}{\begin{tabular}[c]{@{}c@{}}sr5\\ 2684\end{tabular}} &
  \multicolumn{2}{c|}{\begin{tabular}[c]{@{}c@{}}sr10\\ 1900\end{tabular}} &
  \multicolumn{2}{c|}{\begin{tabular}[c]{@{}c@{}}sr15\\ 10552\end{tabular}} &
  \multicolumn{2}{c}{\begin{tabular}[c]{@{}c@{}}sr20\\ 2043\end{tabular}} \\ \midrule
\multicolumn{1}{c|}{\begin{tabular}[c]{@{}c@{}}light\\ 1964\end{tabular}} &
  \multicolumn{1}{c|}{\begin{tabular}[c]{@{}c@{}}dark\\ 0\end{tabular}} &
  \multicolumn{1}{c|}{\begin{tabular}[c]{@{}c@{}}light\\ 1366\end{tabular}} &
  \multicolumn{1}{c|}{\begin{tabular}[c]{@{}c@{}}dark\\ 0\end{tabular}} &
  \multicolumn{1}{c|}{\begin{tabular}[c]{@{}c@{}}light\\ 3004\end{tabular}} &
  \multicolumn{1}{c|}{\begin{tabular}[c]{@{}c@{}}dark\\ 1818\end{tabular}} &
  \multicolumn{1}{c|}{\begin{tabular}[c]{@{}c@{}}light\\ 861\end{tabular}} &
  \begin{tabular}[c]{@{}c@{}}dark\\ 285\end{tabular} &
  \multicolumn{1}{c|}{\begin{tabular}[c]{@{}c@{}}light\\ 2684\end{tabular}} &
  \multicolumn{1}{c|}{\begin{tabular}[c]{@{}c@{}}dark\\ 0\end{tabular}} &
  \multicolumn{1}{c|}{\begin{tabular}[c]{@{}c@{}}light\\ 1900\end{tabular}} &
  \multicolumn{1}{c|}{\begin{tabular}[c]{@{}c@{}}dark\\ 0\end{tabular}} &
  \multicolumn{1}{c|}{\begin{tabular}[c]{@{}c@{}}light\\ 6837\end{tabular}} &
  \multicolumn{1}{c|}{\begin{tabular}[c]{@{}c@{}}dark\\ 3715\end{tabular}} &
  \multicolumn{1}{c|}{\begin{tabular}[c]{@{}c@{}}light\\ 1465\end{tabular}} &
  \begin{tabular}[c]{@{}c@{}}dark\\ 578\end{tabular} \\ \bottomrule
\end{tabular}

}
\label{tab:sr_scene}
\end{table}
\begin{table}[t]
\caption{Distribution of the number of objects in scenes with object scales.}
\centering
\resizebox{\linewidth}{!}{

\begin{tabular}{@{}cccccc|cccccc@{}}
\toprule
\multicolumn{6}{c|}{\textbf{RGB}} &
  \multicolumn{6}{c}{\textbf{Sonar}} \\ \midrule
\multicolumn{3}{c|}{\begin{tabular}[c]{@{}c@{}}light\\ 7195\end{tabular}} &
  \multicolumn{3}{c|}{\begin{tabular}[c]{@{}c@{}}dark\\ 2103\end{tabular}} &
  \multicolumn{3}{c|}{\begin{tabular}[c]{@{}c@{}}light\\ 12886\end{tabular}} &
  \multicolumn{3}{c}{\begin{tabular}[c]{@{}c@{}}dark\\ 4293\end{tabular}} \\ \midrule
\multicolumn{1}{c|}{\begin{tabular}[c]{@{}c@{}}small\\ 197\end{tabular}} &
  \multicolumn{1}{c|}{\begin{tabular}[c]{@{}c@{}}medium\\ 3910\end{tabular}} &
  \multicolumn{1}{c|}{\begin{tabular}[c]{@{}c@{}}large\\ 3088\end{tabular}} &
  \multicolumn{1}{c|}{\begin{tabular}[c]{@{}c@{}}small\\ 0\end{tabular}} &
  \multicolumn{1}{c|}{\begin{tabular}[c]{@{}c@{}}medium\\ 1086\end{tabular}} &
  \begin{tabular}[c]{@{}c@{}}large\\ 1017\end{tabular} &
  \multicolumn{1}{c|}{\begin{tabular}[c]{@{}c@{}}small\\ 7763\end{tabular}} &
  \multicolumn{1}{c|}{\begin{tabular}[c]{@{}c@{}}medium\\ 4195\end{tabular}} &
  \multicolumn{1}{c|}{\begin{tabular}[c]{@{}c@{}}large\\ 928\end{tabular}} &
  \multicolumn{1}{c|}{\begin{tabular}[c]{@{}c@{}}small\\ 2991\end{tabular}} &
  \multicolumn{1}{c|}{\begin{tabular}[c]{@{}c@{}}medium\\ 1302\end{tabular}} &
  \begin{tabular}[c]{@{}c@{}}large\\ 0\end{tabular} \\ \bottomrule
\end{tabular}

}
\label{tab:scene_scale}
\end{table}
\begin{table}[t]
\caption{Distribution of the number of objects in scenes with maximum sonar ranges.}
\centering
\resizebox{\linewidth}{!}{

\begin{tabular}{@{}cccccccc|cccccccc@{}}
\toprule
\multicolumn{8}{c|}{\textbf{RGB}} &
  \multicolumn{8}{c}{\textbf{Sonar}} \\ \midrule
\multicolumn{4}{c|}{\begin{tabular}[c]{@{}c@{}}light\\ 7195\end{tabular}} &
  \multicolumn{4}{c|}{\begin{tabular}[c]{@{}c@{}}dark\\ 2103\end{tabular}} &
  \multicolumn{4}{c|}{\begin{tabular}[c]{@{}c@{}}light\\ 12886\end{tabular}} &
  \multicolumn{4}{c}{\begin{tabular}[c]{@{}c@{}}dark\\ 4293\end{tabular}} \\ \midrule
\multicolumn{1}{c|}{\begin{tabular}[c]{@{}c@{}}sr5\\ 1964\end{tabular}} &
  \multicolumn{1}{c|}{\begin{tabular}[c]{@{}c@{}}sr10\\ 1366\end{tabular}} &
  \multicolumn{1}{c|}{\begin{tabular}[c]{@{}c@{}}sr15\\ 3004\end{tabular}} &
  \multicolumn{1}{c|}{\begin{tabular}[c]{@{}c@{}}sr20\\ 861\end{tabular}} &
  \multicolumn{1}{c|}{\begin{tabular}[c]{@{}c@{}}sr5\\ 0\end{tabular}} &
  \multicolumn{1}{c|}{\begin{tabular}[c]{@{}c@{}}sr10\\ 0\end{tabular}} &
  \multicolumn{1}{c|}{\begin{tabular}[c]{@{}c@{}}sr15\\ 1818\end{tabular}} &
  \begin{tabular}[c]{@{}c@{}}sr20\\ 285\end{tabular} &
  \multicolumn{1}{c|}{\begin{tabular}[c]{@{}c@{}}sr5\\ 2684\end{tabular}} &
  \multicolumn{1}{c|}{\begin{tabular}[c]{@{}c@{}}sr10\\ 1900\end{tabular}} &
  \multicolumn{1}{c|}{\begin{tabular}[c]{@{}c@{}}sr15\\ 6837\end{tabular}} &
  \multicolumn{1}{c|}{\begin{tabular}[c]{@{}c@{}}sr20\\ 1465\end{tabular}} &
  \multicolumn{1}{c|}{\begin{tabular}[c]{@{}c@{}}sr5\\ 0\end{tabular}} &
  \multicolumn{1}{c|}{\begin{tabular}[c]{@{}c@{}}sr10\\ 0\end{tabular}} &
  \multicolumn{1}{c|}{\begin{tabular}[c]{@{}c@{}}sr15\\ 3715\end{tabular}} &
  \begin{tabular}[c]{@{}c@{}}sr20\\ 578\end{tabular} \\ \bottomrule
\end{tabular}

}
\label{tab:scene_sr}
\end{table}

\end{document}